\documentclass[a4paper,fleqn]{cas-sc}
\usepackage[utf8]{inputenc}
\usepackage{textcomp}
\usepackage{tcolorbox}
\usepackage[numbers,sort&compress]{natbib}
\usepackage{float}
\usepackage{pdflscape}
\restylefloat{figure}
\usepackage{caption}
\usepackage{listings}
\usepackage{xcolor}
\usepackage{geometry} 

\definecolor{gray}{rgb}{0.5,0.5,0.5}

\begin{document}
\let\WriteBookmarks\relax

\shorttitle{Harness-Engineered LLM Agents for Crystal Plasticity Workflows}

\shortauthors{S. O. Alfred et~al.}

\title[mode=title]{CP-Agent: A Harness-Engineered Agent for Crystal Plasticity Simulation Workflows}

\author[1]{Samuel Onimpa Alfred}[orcid=0009-0000-2142-0555]

\author[1]{Abhishek Kumar}

\author[1]{Veera Sundararaghavan}[orcid=0000-0002-1213-7958]
\cormark[1]

\affiliation[1]{organization={Department of Aerospace Engineering, University of Michigan},
            city={Ann Arbor},
            postcode={48109},
            state={MI},
            country={USA}}

\cortext[1]{Corresponding author. E-mail: veeras@umich.edu}

\begin{abstract}
Crystal plasticity (CP) simulations are widely used to predict the mechanical behavior of polycrystalline metals, yet their routine application remains hindered by the substantial manual effort of configuring heterogeneous tools, orchestrating multi-step data pipelines, and calibrating constitutive parameters against experimental measurements. These bottlenecks impede productivity in systematic parameter studies, motivating growing interest in automated workflow solutions.

This study presents \textbf{CP-Agent}, a harness-engineered LLM-based agent that autonomously executes complete CP modeling workflows from tasks specified in natural language. Operating under the \textit{ReAct} paradigm, the agent reasons about tool selection and sequencing while delegating numerical search to established optimizers. The harness comprises a minimal system prompt, typed tool definitions, a dispatcher, and a safety-bounded iteration loop, encoding domain knowledge through tool schemas rather than hard-coded logic.

CP-Agent is demonstrated on four case studies: calibrating four slip parameters of additively manufactured stainless steel 316L against tensile data; validating the workflow against published copper benchmarks, reproducing stress-strain and texture evolution; recovering the initial crystallographic texture of copper, where the agent correctly identifies a diffuse initial texture; and reproducing the multi-pass rolling texture evolution of a Mg-Zn-Ca alloy, where the agent chains five deformation passes and recovers the experimentally observed weakened, split basal texture. In all cases, the agent inferred the correct execution sequence from the task statement, robustly across repeated runs, and delivered physically interpretable results. This work establishes harness engineering as a systematic approach to automating CP modeling workflows while maintaining physical interpretability and auditability through visible reasoning traces.
\end{abstract}

\begin{highlights}
\item CP-Agent autonomously runs crystal plasticity workflows from natural-language.
\item The agent infers correct tool sequencing without hard-coded workflows.
\item CP-Agent generalizes across case studies by changing only the tool set. 
\item Four cases: calibration, forward validation, inverse recovery, multi-pass.
\item Visible reasoning traces keep results interpretable and auditable.
\end{highlights}


\maketitle

\section{Introduction}\label{sec:intro}

The mechanical properties of polycrystalline metallic alloys emerge from physical phenomena that span many orders of magnitude in length and time scale. Strength and ductility are governed by dislocation motion at the nanometer scale, grain boundary interactions at the micrometer scale, and the crystallographic texture of the polycrystal aggregate at the millimeter scale~\cite{ref1,ref2}. Crystal plasticity modeling workflows occupies the critical mesoscale link in this hierarchy. It resolves deformation at the level of individual grains, governed by slip on specific crystallographic systems, and couples the grain-level response to a macroscopic boundary value problem. This method predicts the macroscopic mechanical response alongside the evolution of microstructural state, including grain-resolved stress and strain fields, lattice rotation, texture evolution, anisotropy, and strain localization~\cite{ref3,ref4}. Across a broad class of polycrystalline metals, CP modeling delivers quantitative predictions of crystallographic texture evolution, strain hardening anisotropy, yield surface development, and localized deformation. These capabilities distinguish CP from data-driven surrogate models, which cannot easily replicate its explicit resolution of grain-level stress and orientation states~\cite{ref5,ref6}. CP modeling finds application in metal forming, where it predicts material flow and failure during shaping operations such as rolling and deep drawing, as well as in the analysis of anelastic behavior during unloading and the assessment of fatigue and damage.

A persistent bottleneck in this multiscale pipeline is constitutive parameter calibration. Direct prediction of macroscale plasticity from first-principles or dislocation-dynamics models is computationally intractable for engineering-scale problems. The standard practice instead adopts a phenomenological hardening law at the slip-system level and calibrates its parameters against experimental stress-strain data~\cite{ref7,ref8}. Phenomenological slip-system hardening laws~\cite{ref9,ref10}, require the calibration of several parameters governing initial slip resistance, hardening rate, saturation stress, and rate sensitivity. This calibration relies on iterative finite element simulations, guided by expert judgment at each cycle~\cite{ref9,ref10}. Beyond calibration, a complete texture evolution study requires microstructure generation, input file conversion, finite element simulation, orientation extraction, and crystallographic post-processing. Each step in a CP modeling workflow depends on different tools with incompatible interfaces, requiring manual coordination by a domain expert. This reliance on human intervention restricts the method to a small community of specialists and severely limits the throughput of systematic parameter studies. Automating these multi-step workflows would therefore be highly desirable.

Conventional automation in the form of fixed scripts or workflows can handle routine tasks, but it falls short when tools change, data formats vary, or decisions require contextual judgment. What is truly needed are \emph{intelligent} agents: systems that can perceive goals, plan actions, invoke tools, interpret results, and adapt when things go wrong. Large language models (LLMs), generative AI systems trained on vast collections of web text, scientific papers, textbooks, and technical documents, represent a promising pathway toward truly adaptive automation. However, an LLM alone is not an agent; it generates text and machine-readable instructions, but it does not act. By equipping LLMs with the ability to call tools, maintain context, and iteratively refine their outputs based on feedback, they can be elevated from passive predictors to active executors. These LLM-based agents combine the reasoning capabilities of the LLM with the autonomy required to execute complex, multi-step workflows. This makes them uniquely suited to CP modeling workflows, where heterogeneous tools must be chained together and decisions depend on physical insight encoded in text and numerical outputs. Unlike rigid scripts, LLM-based agents can adapt to variations in input data, tool versions, or user objectives, enabling truly flexible and scalable automation.

LLM-based agents have seen growing application across scientific and engineering domains, including materials science~\cite{ref11,ref12,ref13,ref14,ref15,ref16,ref17}, manufacturing process optimization~\cite{ref18,ref19,ref20}, finite element analysis for structural design~\cite{ref21,ref22,ref23,ref24,ref25,ref26}, and autonomous model fitting~\cite{ref27}. In materials science, their primary use has been literature data mining. Ansari and Moosavi introduced Eunomia, a zero-shot autonomous agent that matches fine-tuned model performance across several extraction benchmarks~\cite{ref13}. Dagdelen et al. employed fine-tuned LLMs for structured data extraction across multiple chemistry tasks~\cite{ref14}. Polak and Morgan achieved near-90\% precision and recall using prompt-based conversational models~\cite{ref15}. Gupta et al. extracted over one million polymer-property records from approximately 680,000 full-text articles~\cite{ref16}, while Ghosh and Tewari developed a closed-loop multi-agent system with internal consistency checks to extract thermoelectric and structural data from roughly 10,000 articles~\cite{ref17}.

Beyond literature mining, LLM-based agents have been integrated into engineering design, manufacturing, and simulation workflows. Zeng et al. demonstrated a multi-agent framework for process optimization achieving a 31-fold reduction in wall time versus grid search~\cite{ref19}. Guo et al. proposed an MDO Agent that coordinates parametric modeling, finite element analysis, and structural optimization through LLM-driven tool calls~\cite{ref21}. Xiaorui et al. developed a multi-agent system for autonomous 3D CAD generation from natural language~\cite{ref20}. Shi et al. presented Aethorix v1.0, an agent framework for cement manufacturing process optimization with industrial integration~\cite{ref18}. In computational mechanics, several studies have extended LLM-based agents to finite element analysis. Lin et al. introduced TO-Master, a conversational agent framework that orchestrates topology optimization workflows from natural language, handling geometry, meshing, boundary conditions, solver setup, and postprocessing~\cite{ref24}. In related work, Lin et al. developed CAX-Agent, a lightweight harness for MAPDL simulation automation featuring a recovery ladder that escalates from rule-based patching to model-driven regeneration and human intervention, achieving a 0.9267 completion rate on 50 structural benchmarks~\cite{ref25}. Wilke presented a solver-agnostic framework in which coordinated agents execute the complete computational mechanics pipeline, from perceptual data through geometry extraction, material inference, discretization, solver execution, uncertainty quantification, and code-compliant assessment~\cite{ref26}. Deotale et al. proposed ALL-FEM, an autonomous system integrating agentic AI with fine-tuned LLMs for FEniCS code generation, achieving 71.79\% code-level success on 39 benchmarks across solid, fluid, and multiphysics applications~\cite{ref23}. Alfred and Sundararaghavan introduced an LLM-based agent for end-to-end, data-driven materials theory development~\cite{ref27}. The agent selects equation forms, generates and executes code, and evaluates model fit to data without human intervention. For well-established relationships such as the Hall--Petch equation and Paris law, the agent correctly identifies the governing equation and makes reliable predictions. For specialized relationships such as Kuhn's equation for the HOMO--LUMO gap, performance depends more strongly on the underlying model. The agent can also suggest new predictive relationships, demonstrated through a strain-dependent law for changes in the HOMO--LUMO gap. That study is among the first to demonstrate LLM-driven scientific model discovery from experimental data, yet it addresses the discovery of empirical relationships rather than the calibration of an existing physics-based constitutive model against experimental measurements.

The use of LLM-based agents in multiscale modeling is gaining traction. Zhang et al. developed MooseAgent, a multi-agent framework that automates MOOSE FEM simulations by decomposing natural-language problem statements into solver configuration, input-file synthesis, and error-driven refinement, exploiting MOOSE's multiphysics and multiscale coupling capabilities~\cite{ref28}. Wang et al. proposed DREAMS, a hierarchical agent engine for density functional theory that pairs a planner with domain-specialized worker agents responsible for atomistic structure generation, systematic convergence testing, input preparation, and property extraction, thereby automating the electronic-scale calculations that supply parameters to coarser models~\cite{ref29}. At the atomistic scale, Shi et al. fine-tuned an LLM-based molecular dynamics agent (MDAgent) for LAMMPS script generation, outperforming general-purpose models on the extraction of thermodynamic material parameters~\cite{ref30}. Bridging scales explicitly, Wang et al. introduced Material Buddy (Matty), an AI-agent system that autonomously constructs and executes simulation workflows by coordinating heterogeneous computational codes, multiscale physical models, machine-learning surrogates, and materials databases~\cite{ref31}, while modular multi-agent frameworks such as MatSciAgent route user queries to task-specific agents spanning database retrieval, crystal structure generation, molecular dynamics, and continuum simulation~\cite{ref32}. Similar orchestration strategies have been demonstrated for computational fluid dynamics in Foam-Agent 2.0, an end-to-end composable multi-agent framework for OpenFOAM case setup and execution~\cite{ref33}, and for planner-executor scheduling of high-throughput materials screening campaigns on leadership-class HPC systems~\cite{ref34}. Most directly relevant to the present work, an AI-agent case study on CP finite element simulations reports that a multi-agent system substantially reduces the effort required to learn, configure, and execute CP workflows, compressing tasks that conventionally demand months of manual work into hours of autonomous execution~\cite{ref35}. That study uses a multi-agent decomposition to execute CP workflows for single crystals but falls short of end--to--end integration including polycrystalline mesh instantiation and postprocessing of textures. However, collectively, these efforts establish that LLM-based agents can drive individual simulation codes across scales.

What remains largely unaddressed, however, is the autonomous, end-to-end execution of complete CP modeling workflows across different problem types, including calibration, forward validation, inverse recovery, and chained multi-step deformation, with physical admissibility enforced and each decision left auditable. This requires the agent not merely to run a solver but to reason about prerequisites, convergence, and physical admissibility across the successive steps of each workflow, whether those steps are optimization iterations, chained deformation passes, or a forward run checked against a benchmark.  Two limitations recur across the literature. The first is silent failure. Alfred and Sundararaghavan observed that their LLM-based agents could return incorrect or internally inconsistent equations even when the numerical fit appeared strong~\cite{ref27}, and the same failure mode is implicit in the completion-rate metrics reported elsewhere: CAX-Agent's 0.9267 completion rate~\cite{ref25} and ALL-FEM's 71.79\% code-level success~\cite{ref23} measure whether a workflow terminated or whether code executed, not whether the resulting solution was physically admissible. A CP model calibration is also vulnerable to these issues. An optimizer can drive the objective function down by exploiting non-physical parameter combinations, divergent increments, or an over-refined fit to a single loading path, all of which produce a numerically satisfying result but a physically meaningless model. The second limitation is opacity. When an agent chains dozens of tool calls across meshing, solver execution, and parameter updates, a single final answer offers no basis for judging whether the intermediate reasoning was sound. A plausible-looking parameter set gives the user no way to distinguish genuine physical inference from a coincidence of the optimizer.

These limitations lead to two design choices in this study. First, the \textit{ReAct} paradigm~\cite{ref36} couples reasoning with tool use. During calibration, the agent must justify every action, whether a parameter update or a tool call, with a rationale tied to the task objective. The resulting trace is fully auditable: users can see where the agent's reasoning diverged from simulation evidence, not just inspect the final parameters. Second, the primary engineering effort focuses on the scaffolding around the model rather than on the model itself. Regardless of which LLM is used, the scaffolding ensures consistent behavior, enforces safety, and provides a stable tool interface, while the underlying model can be swapped without disrupting the overall architecture. This focus on scaffolding reflects the principles of harness engineering~\cite{ref37,ref38,ref39,ref40}, an emerging discipline that treats the surrounding infrastructure as the key to agent reliability. A harness is the complete set of scaffolding built around an LLM, including system prompts, tool definitions, orchestration logic, context policies, and feedback mechanisms. This scaffolding transforms an LLM into an autonomous executor capable of completing real-world tasks. The central observation of harness engineering is that a well-engineered harness around a capable LLM often outperforms a more capable LLM with a poorly designed harness. From this perspective, the relevant design question is not which LLM to use, but how to design the scaffolding that allows the LLM to reliably complete the target workflow. In the context of CP model calibration, the harness is where physical admissibility is enforced. Parameter bounds, convergence checks, and multi-path consistency tests are encoded as tool-level constraints and validation gates rather than left to the LLM's discretion. In doing so, the failure modes identified above are structurally prevented rather than merely discouraged by prompting.

To the best of our knowledge, no prior work has applied systematic harness engineering to CP modeling workflows. "Systematic" here refers to designing tool descriptions, system prompt, and agent loop as a coherent, iteratively refined framework that generalizes to structurally different problems without modification. The harness encodes domain knowledge via tool descriptions, with each tool's prerequisites, default parameters, and outputs specified explicitly. The agent infers the correct execution sequence from these descriptions, avoiding hard-coded orchestration logic. The harness is problem-agnostic: new CP modeling workflows require only new tools and descriptions, not changes to the system prompt, agent loop, or orchestration code. Such a demonstration is absent from the existing CP simulation literature.

In contrast to the multi-agent decomposition of \cite{ref35}, the present work instead employs a single, auditable \textit{ReAct} agent and, as developed below, targets closed-loop calibration against experimental data with one harness reused unmodified across forward, calibration, and inverse tasks.Operating under the \textit{ReAct} paradigm~\cite{ref36}, the agent reasons and invokes tools iteratively, with each result fed back into the loop. The harness is built from a minimal system prompt, tool definitions that encode prerequisites and defaults, a dispatcher, and a safety-bounded iteration limit. The agent executes the entire workflow autonomously, from parameter injection and microstructure generation through CP simulation, result extraction, data processing, and figure assembly. The same harness, given only a revised tool set, generalizes without modification to autonomous texture evolution analysis.

This study makes three main contributions. First, it applies harness engineering to CP modeling workflows, integrating microstructure generation, finite element simulation, and crystallographic post-processing within one LLM-driven loop. Second, the harness proves problem-agnostic in practice: the same system prompt, dispatcher, and iteration loop drive all four case studies without modification, with only the tool set and user query changing. Third, it formalizes a tool-description design methodology as the harness's primary knowledge layer, encoding prerequisites, default values, and domain context directly in the tool schemas so the agent sequences the workflow without explicit instructions in the system prompt.

The remainder of this study is organized as follows. Section~\ref{sec:framework} describes the harness engineering framework in general terms, covering architecture, tool description design, system prompt design, simulation infrastructure, and the constitutive model. Section~\ref{sec:case} demonstrates CP-Agent through four case studies: parameter calibration, forward texture validation, inverse texture recovery, and multi-pass texture evolution. All case studies use the same harness, differing only in the tool set and user query. Section~\ref{sec:discussion} discusses implications for multiscale modeling, current limitations, and future directions. Section~\ref{sec:conclusion} presents the conclusions.

\section{Harness-engineered LLM-based Agent Framework}\label{sec:framework}
\begin{figure}[H]
  \centering
  \includegraphics[width=0.75\linewidth]{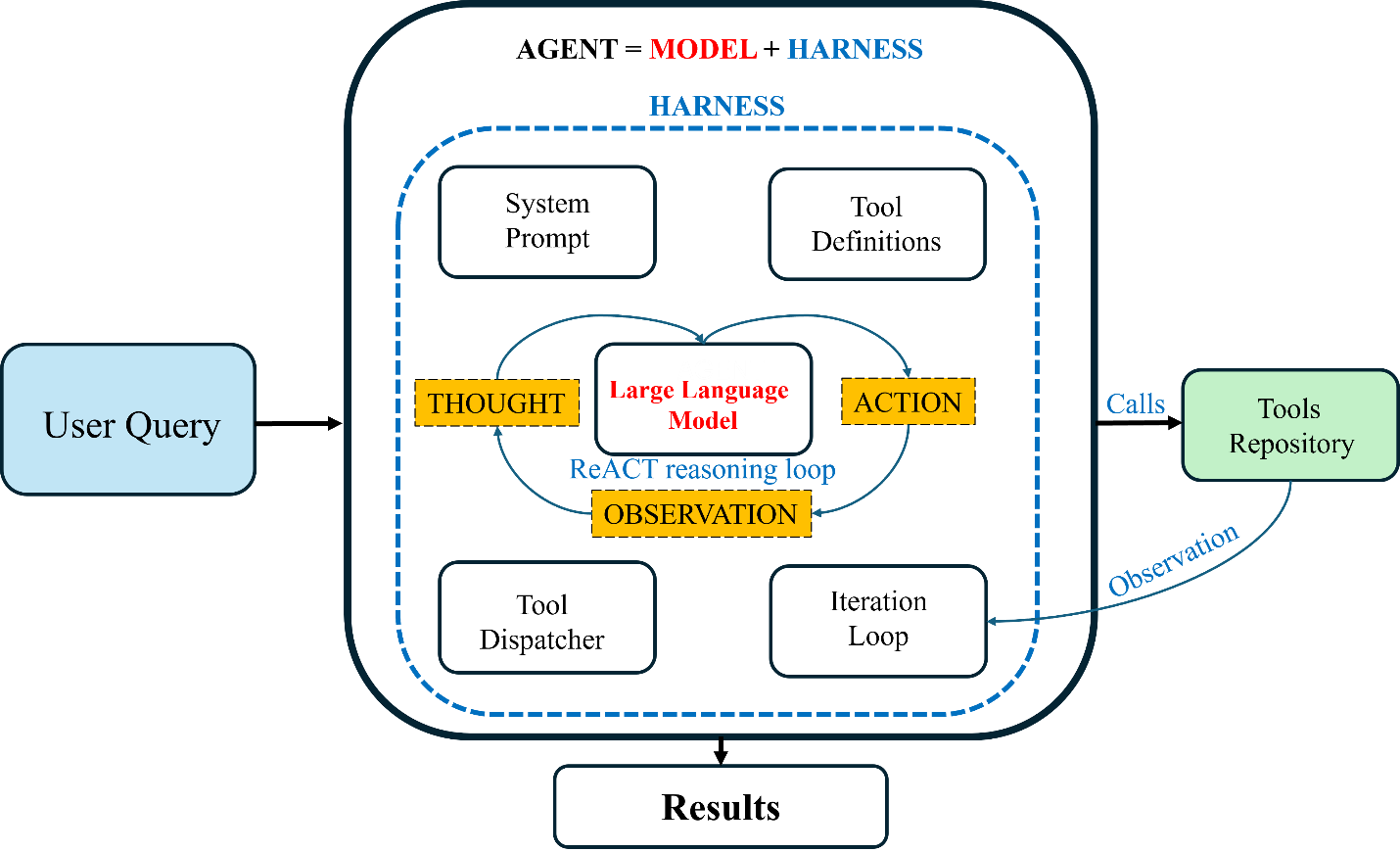}
  \caption{\textit{Architecture of CP-Agent}. The LLM is wrapped by four harness components: the system prompt, the tool definitions, the tool dispatcher, and the iteration loop. The user query enters at runtime; the agent executes a \textit{ReAct} reasoning loop (Thought, Action, Observation) over the tool layer and returns a final response with physical interpretation.}\label{fig:arch}
\end{figure}

This section presents CP-Agent, a harness-engineered LLM-based agent framework (Fig.~\ref{fig:arch}) for CP modeling workflows. The framework is general and problem-agnostic: it is not designed for any specific material or simulation objective, but for the class of workflows in which a user wishes to execute a multi-step CP simulation pipeline spanning microstructure generation, finite element analysis, and crystallographic post-processing by providing a natural language statement of the task. In this study, OpenAI's GPT-4o~\cite{ref41} serves as the underlying LLM, with tools implemented in Python programming language.

Central to the design philosophy adopted for this framework is the distinction between a model and a harness~\cite{ref40}. The model is the underlying LLM, providing reasoning and planning for the task. The harness is the scaffolding around it: the system prompt, tool definitions, dispatcher, and safety-enforcing iteration loop. The harness adapts the model to a domain problem. 

The framework presented in Fig.~\ref{fig:arch} extends the \textit{ReAct}-style single-agent design from our earlier work~\cite{ref27}, where agents performed end-to-end empirical data fitting for materials science relationships. The present harness extends that paradigm in three directions. First, the tool layer provides computationally intensive physics-based simulation codes rather than mathematical curve-fitting routines, requiring the harness to manage long-running sub-processes, multi-format file conversions, and multi-code data pipelines. Second, the same harness is demonstrated across four case studies, with no architectural modification between applications. Third, the concept of semi-autonomous execution is formalized through explicit harness design: the user provides a natural language statement of the task objective, and the harness boundary determines precisely what the agent executes autonomously and what the user must specify.

\subsection{Harness Architecture}\label{sec:arch}
The harness consists of four components, as illustrated in Fig.~\ref{fig:arch}. The \textit{system prompt} is a fixed instruction block loaded at the start of every session, defining the agent's goal and behavioral rules without prescribing an execution sequence. The \textit{tool definitions} are OpenAI function-calling schemas that expose simulation codes as callable functions, with descriptions that embed prerequisites and defaults to guide autonomous sequencing. The \textit{tool dispatcher} receives a tool name and arguments from the model, executes the corresponding code, and returns a structured observation. The \textit{iteration loop} manages the conversation history, sends it to GPT-4o at each step, receives either a tool call or a terminal response, and enforces a hard iteration limit to prevent unbounded execution. Tool invocation uses OpenAI's native function-calling interface, allowing the model to return structured arguments alongside free-text reasoning in a single response.

The architectural design shown in Fig.~\ref{fig:arch} reflects a deliberate division of labor. The LLM's reasoning capacity is spent on decisions where genuine judgment is required, namely the correct sequence of actions, the choice of tools from the repository, and the interpretation of results. It is excluded from the numerical search, where judgment would be a liability or would introduce stochastic variation across multiple runs.
A deliberate design choice is the use of a single agent rather than a multi-agent system. Multi-agent architectures distribute responsibilities across specialized components and introduce inter-agent communication protocols, shared state synchronization, and orchestration overhead that increases harness complexity~\cite{tran2026single}. For the case studies in this study, this complexity is not beneficial. CP modeling workflows are inherently sequential: each tool's output is a required input to the next, and the agent must observe and reason about every intermediate result to make well-reasoned decisions. For example, the agent must confirm that the microstructure was generated correctly before proceeding to the CP simulation step, and must diagnose the source of a simulation failure before deciding whether to retry with modified parameters. These decisions require global context: the full history of tool results must be visible to the reasoning model. A multi-agent system would partition this history across components, risking the loss of cross-step reasoning that enables autonomous error recovery. Moreover, because the workflow is strictly linear, there is no parallelizable sub-task that would benefit from distribution. A single agent with access to the full conversation history provides a simpler, more transparent, and equally capable harness for sequential simulation workflows.

\subsubsection{Harness inputs: system prompt and user query}\label{sec:inputs}
The agent receives two distinct inputs before each execution. The first is the \textit{system prompt}, a fixed harness configuration layer that is automatically loaded at the start of every session without user intervention. The \textit{system promp}t defines the agent's identity, goal, reasoning requirements, and error-handling behavior. It remains invariant across runs and encodes harness-level policies rather than simulation-specific or material-specific knowledge. The second input is the user query, a natural language statement provided at runtime that specifies the task to be performed, including the desired simulation conditions, the calibration objective, or a combination of both. The agent maps the user query to the appropriate tool sequence by reasoning from the tool descriptions.

This separation constitutes a key structural property of the harness. The \textit{system prompt} encodes the agent's operational knowledge and behavioral constraints, including what constitutes correct reasoning, how to handle tool failures, and when to terminate execution. The user query, by contrast, encodes the scientific intent of the specific run. A user can describe the desired simulation in plain language, and the agent translates that description into a precise sequence of tool calls with correct arguments. The harness boundary is thus explicit: everything in the \textit{system prompt} is determined by the harness designer; everything in the user query is determined at runtime by the user. The input prompt overrides the default parameters defined in the tool descriptions, giving the user control over key simulation settings while maintaining the harness's problem-agnostic design.

\subsubsection{Agent loop}\label{sec:loop}
At each iteration, the agent receives the full conversation history, including the \textit{system prompt}, the user query, all prior T\textit{hought-Action-Observation} triplets, and the tool schemas, and produces one of two response types. In the \textit{action }phase, it returns a \textit{Thought}, one to two sentences of visible reasoning expressed as text content, followed by a tool call specifying the tool name and its arguments as a JSON object. The tool is executed by the Python dispatcher, and the result is appended to the conversation history as an observation. In the termination phase, the model returns a text-only response with no tool call, signaling completion. This final response constitutes the agent's interpretation of the simulation results, delivered to the user in natural language. The harness enforces a fixed iteration limit as a safety constraint. This limit serves two purposes: it prevents unbounded autonomous execution in the event of repeated tool failures or non-terminating reasoning loops, and it establishes a clear upper bound on computational cost per run. The number of iterations consumed in a given run depends on the complexity of the workflow and the number of tools invoked; application-specific values are reported in the corresponding case studies.

\subsection{Tool Definitions: The Harness Knowledge Layer}\label{sec:tools}
Each tool is defined with an OpenAI function-calling schema that specifies three things: a name, a natural-language description, and a typed list of parameters. These descriptions are where the harness injects its domain knowledge, and they do two jobs at once. First, they tell the model what each tool does, what inputs it needs, and what it returns. Second, because each description also states what must already be available before a tool can run, the agent can read the descriptions together with the user's task statement and work out the correct order of operations on its own, without being given an explicit sequence. This inferred ordering does not rely on any single cue. It is encoded redundantly in the tool names, in the functional meaning of the descriptions, and in the task statement, so the correct sequence is recovered even when individual cues are weakened. This robustness is examined directly in Section~\ref{sec:harness_robustness}.

The tool set is application-dependent and constitutes the primary extension point of the framework. The core simulation tools and the optimizer tools are described in Tables~\ref{tbl:tools} and~\ref{tbl:opt}, respectively. The tool names, descriptions, and parameter schemas presented here reflect the case studies in Section~\ref{sec:case}. Nevertheless, the underlying design methodology, which includes encoding prerequisites in tool descriptions and specifying defaults in parameter schemas, applies identically to any CP modeling workflow. More details on the tools for each case study are provided in \url{https://github.com/samoalfred/harness-cp-agents}.
\begin{landscape}
\begin{table}[width=\linewidth,cols=3,pos=!ht]
\caption{\textit{Tools available to the agent for calibration and texture evolution workflow}.}\label{tbl:tools}
\begin{tabular*}{\tblwidth}{@{\extracolsep{\fill}}>{\raggedright\arraybackslash}p{0.20\textwidth}>{\raggedright\arraybackslash}p{0.22\textwidth}>{\raggedright\arraybackslash}p{0.52\textwidth}@{}}
\toprule
Tool & Key inputs & Description \\
\midrule
inject\_simulation\_parameters & orientation\_type, fiber\_direction, sigma\_spread, and slip-system hardening parameters & Writes texture and constitutive slip parameters into the microstructure generation script and the CP configuration file via regular-expression substitution. Enforces schema-defined defaults for any parameter not explicitly specified in the user query. \\
generate\_microstructure & None (reads from injected config) & Launches MATLAB in batch mode to generate a synthetic polycrystalline microstructure using MTEX and saves the grain ID map and per-grain Rodrigues orientation vectors to an HDF5 file. Also computes and saves pre-deformation pole figures. \\
convert\_hdf5\_to\_prisms & None (reads HDF5 output) & Reads the HDF5 microstructure file and converts it to the CP code input format. Applies Fortran column-major array ordering to preserve the MATLAB memory layout. Updates the CP configuration file with the correct voxel grid dimensions. \\
run\_simulation & None (reads CP configuration file) & Executes a CP simulation using the current configuration. Streams solver output in real time with a configurable timeout. Produces per-quadrature-point output files containing deformed grain orientations and stress-strain data. \\
extract\_post\_orientations & None (reads CP output) & Locates the final per-quadrature-point output file and extracts the post-deformation Rodrigues vector components. Saves extracted orientations for use in post-deformation texture analysis. \\
generate\_pole\_figures & None (reads pre- and post-deformation orientations) & Runs a MATLAB post-processing script that computes orientation distribution functions (ODFs) for both orientation sets and generates \{100\}, \{110\}, and \{111\} pole figures with matched color scales. \\
create\_comparison\_figure & None (reads pole figure image files) & Assembles pre- and post-deformation pole figure images into a publication-quality comparison figure saved at 300 DPI. \\
compare\_stress\_strain & reference stress-strain curve file & Overlay the simulated axial true stress on the reference curve. \\
\bottomrule
\end{tabular*}
\end{table}

\begin{table}[width=\linewidth,cols=3,pos=!ht]
\caption{\textit{Optimizer tools available to the agent for the constitutive parameter calibration workflow}.}\label{tbl:opt}
\begin{tabular*}{\tblwidth}{@{\extracolsep{\fill}}p{0.22\textwidth}p{0.28\textwidth}p{0.44\textwidth}@{}}
\toprule
Tool & Algorithm & When the agent selects this tool \\
\midrule
run\_bayesian\_optimization & Latin Hypercube Sampling + Gaussian Process + Expected Improvement & Default selection when no specific preference is stated. Most sample-efficient optimizer; preferred when the simulation budget is limited or an initial estimate of the optimum is unavailable. \\
run\_differential\_evolution & Population-based global search with mutation and crossover & Selected when the user specifies a broad or global parameter search, or when the query indicates uncertainty about the location of the optimum in the parameter space. \\
run\_nelder\_mead & Nelder-Mead simplex local optimizer & Selected when the user provides an initial parameter estimate and requests fast local refinement. Converges quickly near the optimum but may fail to escape local minima. \\
run\_random\_bo & Random Search followed by Bayesian Optimization & Selected when the user requests an exploratory phase before exploitation, or when a broad baseline is needed before Gaussian Process refinement. \\
\bottomrule
\end{tabular*}
\end{table}
\end{landscape}

\subsection{System Prompt Design}\label{sec:prompt}
The \textit{system prompt} is the behavioral layer of the harness. A minimal philosophy was adopted: the prompt provides a goal and a tool list but does not prescribe an execution sequence. This design, termed goal-driven prompting~\cite{ref42}, contrasts with guided-workflow prompting, where the system prompt enumerates the steps to be followed. A prompted sequence hard-codes the workflow in natural language; the agent follows because it was told to, not because it reasoned. Such a harness cannot generalize beyond the prescribed sequence and fails silently when conditions deviate. The goal-driven prompt delegates sequencing entirely to the tool descriptions. The agent derives execution order by reading prerequisite statements and reasoning about which calls satisfy them. Domain knowledge resides in the tool schemas, close to the operations they govern, rather than in the system prompt, where it would mix with behavioral policies. A sample of the \textit{system prompt} used in this study is shown below, with domain-specific terms adapted to the target application:

\begin{tcolorbox}[colback=blue!10,colframe=gray!40,boxrule=0.4pt]
\itshape
``You are an autonomous agent for crystal plasticity simulation. Your goal: fulfill the user's request using the tools available to you. [tool list]. Read each tool's description carefully. The descriptions tell you what each tool does, what it requires, and what it produces. Use that information to reason about which tools to call, in what order, and with what arguments. When you have completed the task, stop calling tools and deliver your final response.''
\end{tcolorbox}

All four case studies use the goal-driven ReAct configuration: the agent receives only the task objective and the typed tool schemas, and infers the operation sequence from the prerequisites
declared in each schema. The case studies differ only in their tool sets, not
in the harness or the prompting style.

\subsection{Simulation Infrastructure}\label{sec:infra}
The tool layer of the harness includes three simulation capabilities that together constitute a complete CP modeling workflow: microstructure generation, finite element simulation, and crystallographic post-processing. Each capability is implemented as a call to an existing domain-specific code, wrapped by a Python tool function that handles input formatting, subprocess execution, execution monitoring, and output parsing. The tool functions return plain-text observation strings that the agent reads and reasons about directly. The harness architecture places no constraint on which specific finite element solver, texture analysis package, or microstructure generation tool is used; adapting the framework to alternative codes requires only replacing the corresponding tool implementations and updating their descriptions, with no change to the harness core.

\subsubsection{Microstructure Generation}\label{sec:micro}
A microstructure generation package defines crystallographic orientation distributions, samples grain orientations, and computes derived quantities such as orientation distribution functions (ODFs) and pole figures. Grain orientations are drawn from a fiber texture distribution centered on either a user-specified direction or an experimentally measured EBSD crystallographic direction. The pre‑deformation ODFs and pole figures are computed in the same step. The resulting microstructure is saved as an HDF5 file containing the grain ID map and per‑grain Rodrigues orientation vectors. For automated execution, the package is launched from Python in batch mode, while the tool wrapper handles platform‑specific path conversions.

The microstructures used in this study are generated using the microstructure generator (MicroGen) implemented in the ELAS3D-Xtal package~\cite{ref43}. The generation scripts are written in MATLAB and handle grain orientation sampling, texture definition, and microstructure output.

\subsubsection{Crystal Plasticity Solver}\label{sec:fem}
CP simulation constitutes the computational core of the workflow. The solver takes a discretized polycrystal microstructure and a set of constitutive and loading parameters, solves the crystal plasticity boundary value problem at the grain level, and outputs per-quadrature-point quantities including deformed grain orientations, stress components, and accumulated slip on each slip system. The simulation domain is configured through a plain-text parameter file specifying the voxel grid dimensions, elastic constants, constitutive model parameters, loading conditions, and output frequency. The parameter injection tool modifies this file directly via regular-expression substitution, allowing the agent to configure the simulation for any user-specified set of material parameters without manual file editing. The solver binary is invoked as a subprocess from Python; its output is streamed in real time so the user and agent can monitor progress, and a configurable timeout is enforced as a harness-level safety constraint.

In this study, CP simulations are performed using PRISMS-Plasticity~\cite{ref44}, an open-source, parallel 3D CP software package. Both constitutive models built into the software are used: the rate-independent model of Anand and Kothari~\cite{ref45} for the parameter-calibration case study (CS1), and a rate-dependent power-law flow rule for the texture-evolution, inverse, and multi-pass case studies (CS2, CS3, and CS4). Further details of both formulations are provided in Section S1 of the Supplementary Material.

\subsubsection{Post-deformation Texture Analysis}\label{sec:post}
Post-deformation grain orientations are extracted from the per-quadrature-point output files written by the CP solver. The Rodrigues vector components of the deformed lattice orientation are extracted from the final output file, aggregated by grain, and passed to the texture analysis software for post-deformation ODF computation and pole figure generation. A key feature of the post-processing workflow is the matched color scale between pre- and post-deformation pole figures. For each Miller index family $\{hkl\}$, the maximum ODF intensity across both the pre- and post-deformation distributions is determined, and both pole figures are rendered with an identical color range. Without this matching, texture sharpening or weakening is visually indistinguishable from a rescaling of the color axis, making the comparison physically ambiguous. The matched scale ensures that a higher peak intensity in the post-deformation figure unambiguously indicates texture strengthening. In this study, all texture analyses are performed using MTEX~\cite{ref46}.

\section{Case Studies}\label{sec:case}
This section demonstrates the CP-Agent on four CP modeling workflows, chosen for their contrasting harness demands (summarized in Table~\ref{tbl:cs}). The first, a parameter calibration task (CS1), addresses the problem of known experimental data with unknown material parameters: the agent searches for slip parameters that fit the target stress--strain curves. The second, a forward validation task (CS2), assumes known parameters and known microstructure: the agent executes the simulation and checks that the outputs match published results. The third, an inverse texture task (CS3), starts from a known target texture and unknown initial microstructure: the agent searches for the microstructure that yields the target. The fourth, a multi-pass texture-evolution task (CS4), chains five deformation passes of a magnesium alloy, carrying the deformed microstructure of each pass into the next and validating the texture evolution against experiment.All four workflows use the same LLM and tool dispatcher, differing only in harness configuration.

\begin{table}[!ht]
\centering
\small
\caption{\textit{Case study specifications.} The four workflows differ in their objective, unknowns, and simulation requirements while using the same agent harness with only configuration changes.}
\label{tbl:cs}
\begin{tabular*}{\linewidth}{@{\extracolsep{\fill}}
  >{\raggedright\arraybackslash}p{0.12\linewidth}
  >{\raggedright\arraybackslash}p{0.19\linewidth}
  >{\raggedright\arraybackslash}p{0.19\linewidth}
  >{\raggedright\arraybackslash}p{0.19\linewidth}
  >{\raggedright\arraybackslash}p{0.19\linewidth}@{}}
\toprule
Feature & CS1: Calibration & CS2: Forward Validation & CS3: Inverse Problem & CS4: Multi-pass Evolution \\
\midrule
Material & SS316L & Copper & Copper & ZX31 (Mg--3Zn--0.3Ca), HCP \\
Question & What slip parameters fit the stress-strain data? & Does the model reproduce a known benchmark? & What initial texture produces the target deformation texture? & How does the basal texture evolve over multiple rolling passes? \\
Unknown (solved for) & Four slip parameters ($s_0$, $h_0$, $s_s$, $n$) & none (single forward run) & initial texture (random vs fiber, spread) & none (chained forward run) \\
Held fixed & microstructure, loading & everything (literature params) & slip params, loading, grain count & everything (Table-1 params) \\
Target/reference & experimental tensile curve~\cite{ref48} & experimental compression curve~\cite{yaghoobi2022prisms}; pole figure~\cite{yaghoobi2022prisms} & pole figures~\cite{yaghoobi2022prisms} & 5-pass simulation + EBSD pole figures~\cite{yaghoobi2025effects} \\
Method & agent selects optimizer $\rightarrow$ search & agent runs pipeline once & agent selects Bayesian optimizer $\rightarrow$ search over textures & agent chains five passes \\
Number of simulations & 10--60 (search) & 1 & 15 (search) & 5 (one per pass) \\
Output checked against & Stress-strain curve (RMSE/MAPE) & Stress-strain curve + pole figures & pole-figure loss vs target & (0001) texture evolution + pole figures \\
\bottomrule
\end{tabular*}
\end{table}

\subsection{Computational setup}\label{sec:setup}
All case studies have the same structure: the natural language query provided to the agent, the LLM, which is the reasoning engine, the harness configuration, the execution trace, and the results. All agents are implemented in Python 3.7 running under Windows Subsystem for Linux. The LLM used is GPT-4o, accessed through the OpenAI function-calling interface~\cite{ref47}, which supplies the typed tool schemas described in Section~\ref{sec:framework} and returns tool calls as structured JSON rather than free text. The LLM's sampling temperature is 0.1, and responses are capped at 1024 tokens per turn. This low temperature is deliberate, as it ensures reliability and consistency across multiple runs. The implementation details described below apply uniformly across all case studies.

Microstructures are generated using the microstructure generator in ELAS3D-Xtal~\cite{ref43}. Outputs are written to HDF5 and converted to the solver's native input format by a routine in the tool layer. Finite element simulations are performed with PRISMS-Plasticity~\cite{ref44}, invoked by the harness as an external process. Post-processing, including stress--strain extraction, error metric computation, orientation distribution function estimation, and figure assembly, is performed in Python. The agent never edits solver source code; instead, every tool provides a fixed, validated operation over configuration files and outputs. The run-to-run reliability of the agent and the robustness of its autonomous sequencing are characterized separately in Section~\ref{sec:harness_robustness}, after the case studies.

\subsection{Case Study 1: Parameter Calibration}\label{sec:cs1}
The first case study identifies four slip parameters in the rate-independent crystal plasticity model of Anand and Kothari~\cite{ref45}. These parameters are calibrated to reproduce the experimental uniaxial tensile stress-strain curve for laser powder bed fusion manufactured Stainless Steel (SS)316L, using data from Douglas et al.~\cite{ref48}. This is a canonical inverse problem in crystal plasticity model development. It is computationally expensive because every candidate parameter set requires a full finite element simulation. Consequently, only a limited number of fit error evaluations are feasible within practical time constraints, and each evaluation must be used strategically.

\subsubsection{Task specification}\label{sec:cs1task}
The agent receives the following natural language query:

\begin{tcolorbox}[colback=blue!10,colframe=gray!40,boxrule=0.4pt]
\itshape
``Generate a random SS316L microstructure for crystal plasticity simulations using PRISMS-Plasticity. Then calibrate the crystal plasticity slip parameters for SS316L to match the experimental tensile stress-strain data (SS316L\_experiment.txt). Use Bayesian Optimization. Tune: Initial Slip Resistance [100--150 MPa], Initial Hardening Modulus [800--2500 MPa], Saturation Stress [350--600 MPa], Power Law Exponent [1--3]. Stop when RMSE $<$ 5 MPa or MAPE $<$ 2\% or after 60 simulations.''
\end{tcolorbox}

The elastic constants, other input parameters, and the microstructure details for random microstructure generation in MicroGen are all omitted from the query because they are fixed inputs: the first in the PRISMS-Plasticity input file, and the second in MicroGen. More details on the setup can be found here: \url{https://github.com/samoalfred/harness-cp-agents/tree/main/case_study_1_calibration}

The experimental data consists of a stress versus strain curve for SS316L, sampled at sixteen points across strains ranging from zero to 0.3\%. Agreement between the simulated and experimental curves is quantified using the root mean square error (RMSE) in MPa and the mean absolute percentage error (MAPE). Both metrics are computed after interpolating both curves onto common strain points.

\subsubsection{Harness configuration}\label{sec:cs1config}
The harness provides four optimization algorithms as tools, together with two microstructure preparation tools. Consistent with the goal-driven configuration used throughout, the agent determines the sequence of actions from the tool schemas and interprets the results. A critical design choice is that the agent never proposes parameter values or performs the optimization itself. Each optimizer tool is a self-contained Python routine that manages the entire search: it proposes candidates, calls the simulation wrapper, computes error metrics, logs evaluations, and enforces stopping criteria internally. Each search strategy is a standard implementation (for example, Bayesian optimization, differential evolution, and Nelder--Mead);the model only selects which strategy to run, and that routine performs the numerical search deterministically.

Table~\ref{tbl:search} lists the four slip parameters, their bounds, and the stopping criteria. The stopping criteria are enforced within the simulation wrapper rather than inside the optimizers themselves. The wrapper raises a dedicated convergence exception as soon as any criterion is met, halting the search immediately. This ensures that the simulation budget is never exceeded, regardless of which algorithm the agent selects from Table~\ref{tbl:opt} or how that algorithm's own termination logic behaves.

\begin{table}[width=\linewidth,cols=4,pos=!ht]
\caption{\textit{Search space and stopping criteria, enforced by the harness independently of the query}.}\label{tbl:search}
\begin{tabular*}{\tblwidth}{@{\extracolsep{\fill}}p{0.26\textwidth}p{0.12\textwidth}p{0.14\textwidth}p{0.40\textwidth}@{}}
\toprule
Parameter & Lower & Upper & Physical Meaning and its role \\
\midrule
Initial slip resistance, $s_0$ & 100 MPa & 150 MPa & Stress to initiate slip; controls yield stress \\
Initial hardening modulus, $h_0$ & 800 MPa & 2500 MPa & Rate of initial strengthening; controls initial hardening rate \\
Saturation stress, $s_s$ & 350 MPa & 600 MPa & Maximum achievable stress; controls the flow stress plateau \\
Power law exponent, $n$ & 1 & 3 & Strain rate sensitivity; controls sharpness of the hardening transition / strain rate sensitivity \\
Stopping: RMSE & - & $<$ 5 MPa & Absolute agreement criterion \\
Stopping: MAPE & - & $<$ 2 \% & Relative agreement criterion \\
Stopping: budget & - & 60 simulations & Hard computational cap \\
\bottomrule
\end{tabular*}
\end{table}

\subsubsection{Results}\label{sec:cs1res}
\begin{figure}[H]
  \centering
  \includegraphics[width=\linewidth]{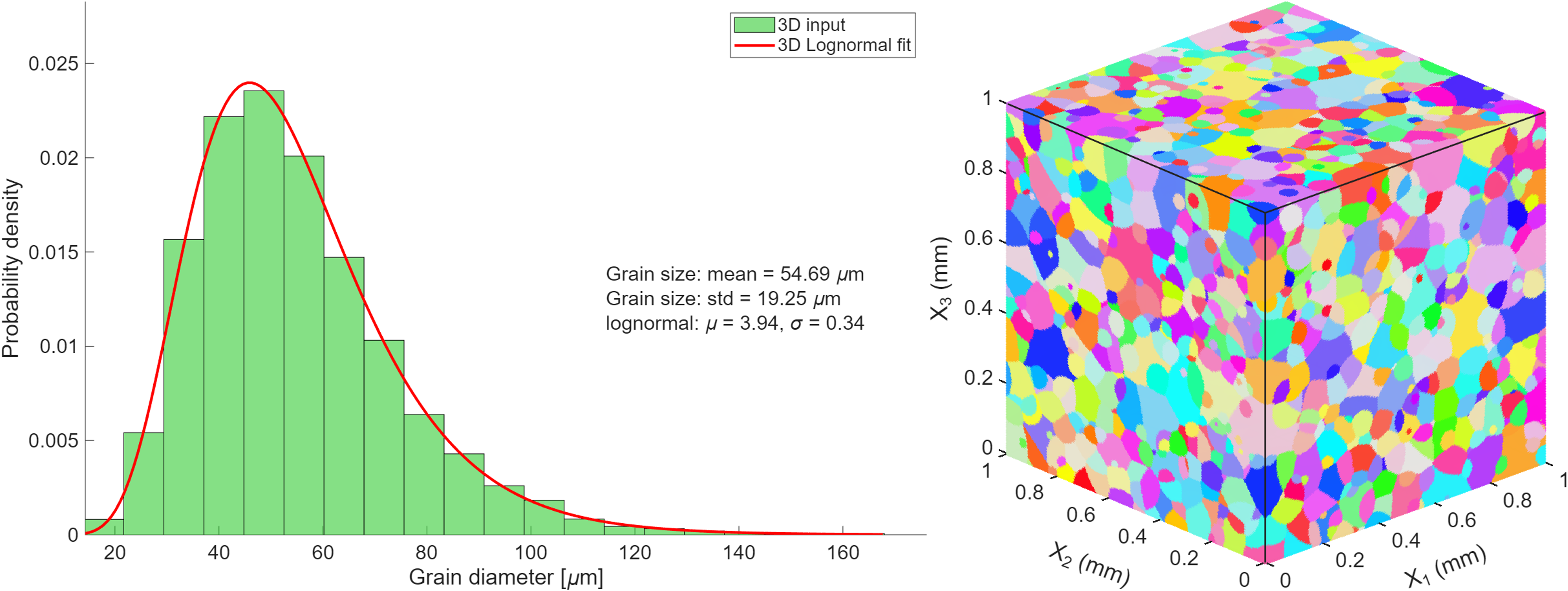}
  \caption{Microstructure generation. (Left) Input grain statistics. (Right) Representative volume element (RVE) generated by the agent.}\label{fig:micro}
\end{figure}

The agent completed the task successfully. It generated the microstructure shown in Fig.~\ref{fig:micro}, converted the grain ID and orientation data into the format required by PRISMS-Plasticity, and recognized the user's explicit request for Bayesian optimization, selecting that optimizer from the tool repository. Control is then passed entirely to the optimizer for the duration of the search. Upon the optimizer's return, the agent received a single structured observation containing the best parameter set and its error metrics, and produced a physical interpretation. The agent's full reasoning trace, detailing each decision step, is provided in Section S2 of the Supplementary Material.

\begin{figure}[H]
  \centering
  \includegraphics[width=0.65\linewidth]{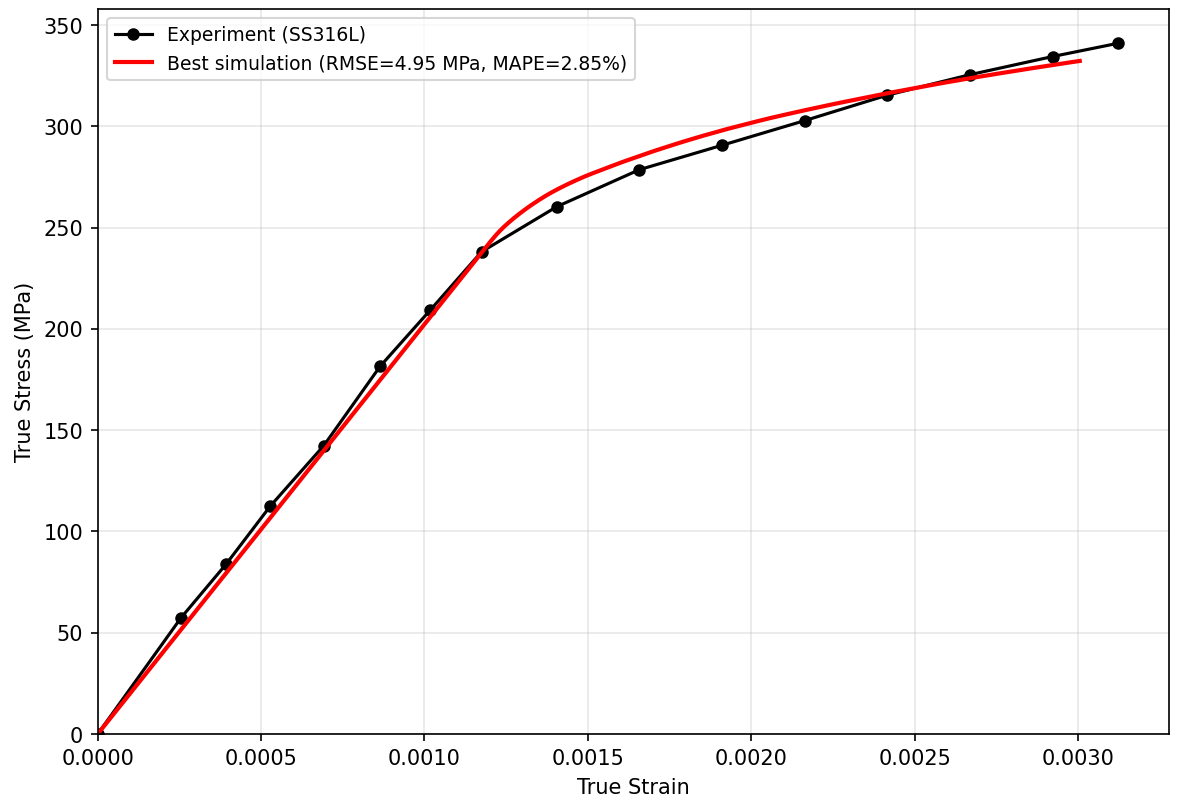}
  \caption{\textit{Best-fit simulated stress-strain response (red solid line) versus experimental tensile data for SS316L (black markers and line)}.}\label{fig:ss1}
\end{figure}

\paragraph{Calibrated parameters and fit quality}\label{sec:cs1calib}
Table~\ref{tbl:calib} shows the autonomously obtained calibrated slip parameters while Fig.~\ref{fig:ss1} shows the corresponding best-fit stress-strain response. The optimization achieved high-fidelity reproduction of the experimental tensile behavior, with an RMSE of 4.95 MPa, below the 5 MPa stopping threshold. The MAPE of 2.85\%, however, remained marginally above the 2\% target. All calibrated parameters fell within their prescribed bounds.

\begin{table}[width=\linewidth,cols=3,pos=!ht]
\caption{Calibrated slip parameters for SS316L obtained by the agent-selected Bayesian optimizer.}\label{tbl:calib}
\begin{tabular*}{\tblwidth}{@{\extracolsep{\fill}}p{0.32\linewidth}p{0.18\linewidth}p{0.40\linewidth}@{}}
\toprule
Quantity & Value & Comment \\
\midrule
Initial slip resistance, $s_0$ & 121 MPa & Interior of the bounds \\
Initial hardening modulus, $h_0$ & 2399 MPa & 97.5 per cent of the upper bound \\
Saturation stress, $s_s$ & 542 MPa & 97.8 per cent of the upper bound \\
Power law exponent, $n$ & 2.9 & Interior of the bounds \\
RMSE & 4.95 MPa & Below the 5 MPa stopping threshold \\
MAPE & 2.85 \% & Above the 2 per cent stopping threshold \\
Terminating evaluation & 26 & Of a 60-simulation budget \\
\bottomrule
\end{tabular*}
\end{table}

\begin{figure}[H]
  \centering
  \includegraphics[width=\linewidth]{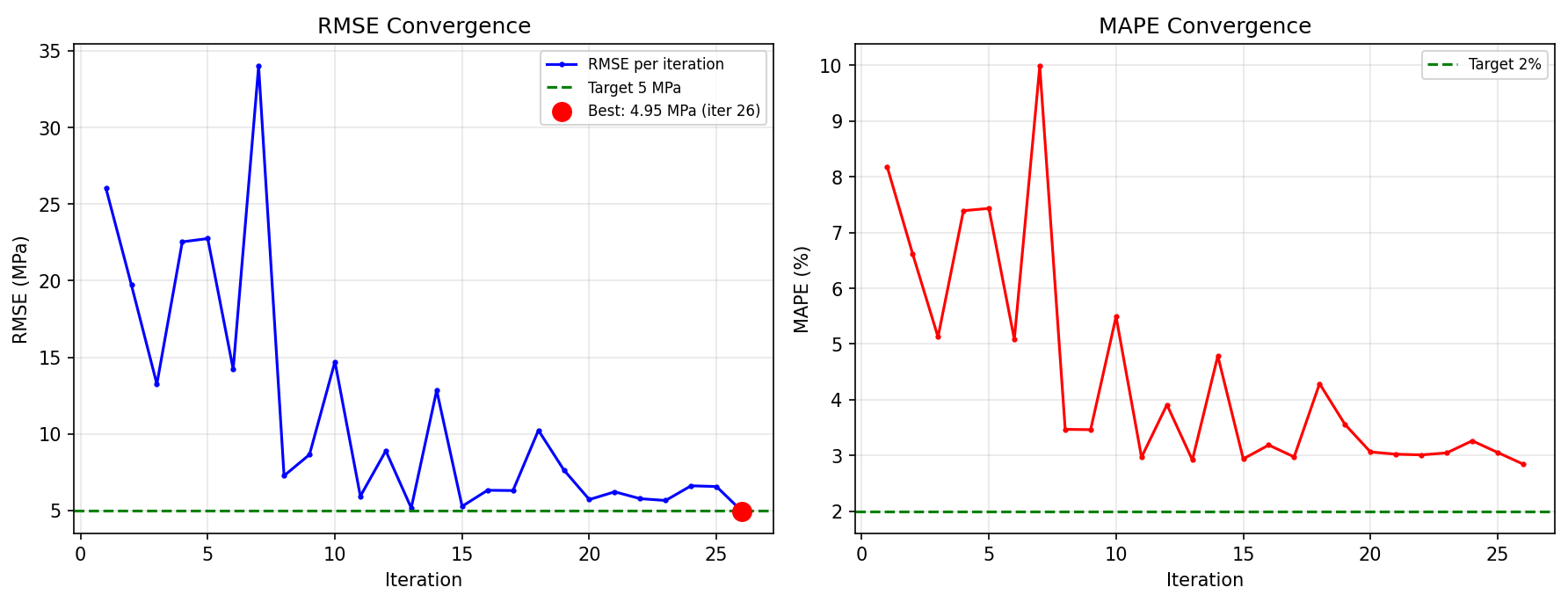}
  \caption{\textit{Convergence history of the agent-selected Bayesian optimization.} Left: RMSE against evaluation number on a logarithmic scale, with the 5 MPa target as a dashed line and the converged point marked. Right: MAPE against evaluation number with the 2\% target.}\label{fig:conv}
\end{figure}

\paragraph{Convergence}\label{sec:cs1conv}
The agent tracked convergence by selecting the appropriate tool from the repository. Convergence behavior is shown in Fig.~\ref{fig:conv}, where stopping thresholds were set at RMSE $<$ 5 MPa or MAPE $<$ 2\%. The optimization met the RMSE criterion at 26th iteration, well within the 60-simulation budget, achieving a final RMSE of 4.95 MPa.

The convergence history shows the expected two-phase behavior. The first ten iterations sample the parameter space broadly and produce errors up to 34 MPa. The large oscillations in this phase are not a defect but the intended behavior of Latin hypercube sampling, which deliberately spreads points across the domain rather than descending. From iteration 11 onward, the amplitude of the oscillation contracts markedly as the Gaussian process concentrates sampling in the low-error region, and the search terminates as soon as the threshold is crossed (below the dashed green line of Fig.~\ref{fig:conv}). The MAPE trace descends in parallel but flattens above its 2\% target, a point returned to below.

\subsubsection{Discussion}\label{sec:cs1disc}
This case study demonstrates, in the calibration setting, the design principle on which the harness rests. The agent is responsible for selecting and sequencing a numerical procedure, not for carrying out the numerical search itself. From a single natural language goal, and without being told the order of operations, the agent generated the microstructure, converted it to solver input, recognized that the query and its own tool schema both pointed to Bayesian optimization for an expensive, budget-limited search, and handed the entire parameter search to that optimizer. 
A calibration workflow follows a natural order, and the harness reserves the LLM's reasoning capacity for decisions that genuinely require judgment: operation sequencing, algorithm selection, and result interpretation. Numerical search, in contrast, is delegated to established optimizers, where the LLM's stochastic nature and lack of physics understanding would be a liability rather than an advantage.

A caveat concerns the identifiability of the calibrated set. 
The experimental curve of Douglas et al.~\cite{ref48} extends to a true strain of only approximately 0.3\%, that is, through the elastic--plastic knee but not into the fully developed flow regime. 
The saturation stress, $s_s$ and initial hardening modulus, $h_0$ principally govern the flow-stress plateau, which this strain window does not reach; consistently, both parameters converged near their upper bounds (Table~\ref{tbl:cs}), whereas $s_0$ and $n$, which control the yield onset and the sharpness of the transition and are well constrained by the available data, settled in the interior. 
The recovered $s_0$ and $n$ should therefore be read as physically identified, while $s_s$ and $h_0$ are only loosely constrained and partly compensating over this window. 
This is a property of the target data, not of the agent or the optimizer: the harness faithfully minimized the stated objective. 
It also illustrates, in a concrete setting, the silent-failure risk raised in the Introduction, and motivates calibrating against a curve that reaches the flow plateau, or against multiple loading paths, when the plateau parameters are of interest.

\subsection{Case Study 2: Texture Evolution}\label{sec:cs2}

The first case study demonstrated the agent's ability to calibrate slip parameters against a target curve. The second case study shifts to a forward-prediction task: the agent is given fixed literature parameters for oxygen-free high-conductivity (OFHC) copper, specifically elastic constants $C_{11} = 170$ GPa, $C_{12} = 124$ GPa, $C_{44} = 75$ GPa; slip parameters $s_0 = 16$ MPa, $h_0 = 200$ MPa, $s_s = 129.5$ MPa, $n = 2$; and a latent-hardening ratio $q = 1.4$, together with a rate-dependent flow rule (reference shearing rate $1\times10^{-3}$ s$^{-1}$ and rate-sensitivity exponent $m = 77$). Its objective is to generate a random-texture polycrystal, compress it under a Taylor-homogenized crystal-plasticity model, and report both the mechanical response and the resulting deformation texture. The outputs, specifically the true stress--strain curve and the deformed pole figures, are validated against the PRISMS-Plasticity TM results of Yaghoobi et al.~\cite{yaghoobi2022prisms}, who in turn benchmarked the experimental measurements of Bronkhorst et al.~\cite{bronkhorst1992polycrystalline} and the crystal-plasticity simulation of Anand and Kothari~\cite{ref45}. No parameters are tuned; this run is a single autonomous forward pass.

\subsubsection{Task specification}\label{sec:cs2task}
The agent receives the following input query:

\begin{tcolorbox}[colback=blue!10,colframe=gray!40,boxrule=0.4pt]
\itshape
``Perform a forward validation of polycrystalline OFHC copper using the rate-dependent Taylor crystal-plasticity model. First generate a synthetic copper polycrystal of approximately 400 equiaxed grains with a random initial texture, then run a single simulation of uniaxial compression along $Z$ through an imposed velocity gradient to a true strain of about 1.0 (100\% compression), using the fixed slip parameters in prm.prm. Then perform two comparisons. First, the crystallographic texture: plot the $\{111\}$, $\{100\}$, and $\{110\}$ pole figures ($Z$ at centre) on a fixed 0--3.5 MRD scale, producing both the pre- versus post-deformation figure and an aligned comparison of the post-deformation pole
figures against the reference pole figures; the deformed texture should reproduce the $\langle 110\rangle$ compression fiber. Second, compare the simulated von Mises equivalent stress against the reference stress-strain curve over the overlapping strain range, reporting RMSE and MAPE. Provide a brief physical interpretation of the texture evolution (the development of the $\langle 110\rangle$ compression fiber) and of the stress-strain agreement for copper.''
\end{tcolorbox}

The elastic, slip, and rate parameters are omitted from the query because they are set as fixed inputs in the PRISMS-Plasticity input file. More details on the setup can be found here: \url{https://github.com/samoalfred/harness-cp-agents/tree/main/case_study_2_forward}

\subsubsection{Results}\label{sec:cs2res}
From the input query and tool descriptions alone, the agent recovered the correct multi-step workflow shown in Table~\ref{tbl:tools}: generate the microstructure, convert it to PRISMS-Plasticity input, run the simulation, extract the post-deformation orientations, generate the pole figures, assemble the texture comparison, and compare the stress--strain response. This ordering was not prescribed in the prompt; it was inferred from the prerequisites declared in each tool schema. The full reasoning trace is provided in the Supplementary Material.

\begin{figure}[H]
  \centering
  \includegraphics[width=0.65\linewidth]{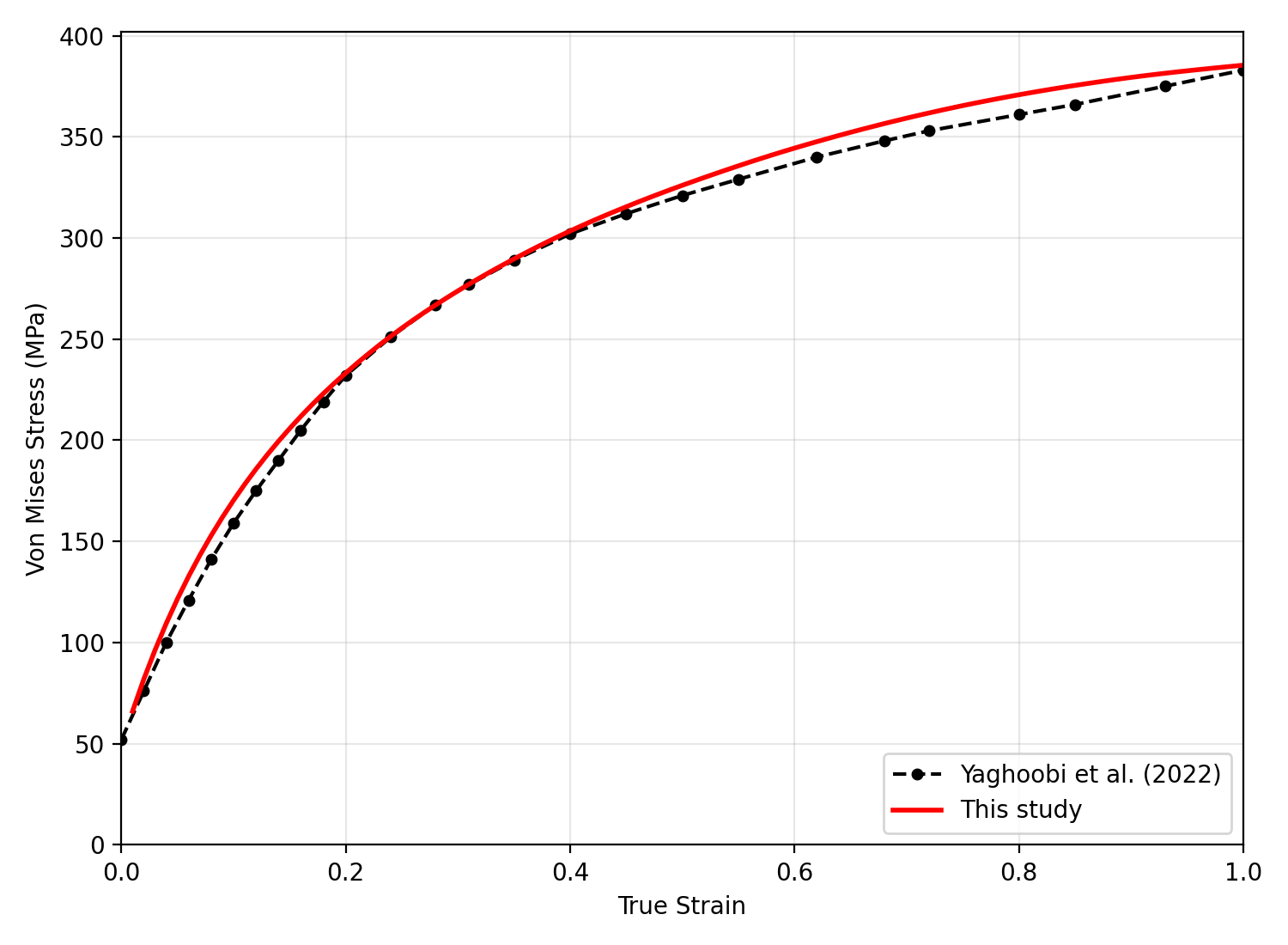}
  \caption{Simulated von Mises equivalent stress versus true strain, compared with the PRISMS-Plasticity TM reference curve of Yaghoobi et al.~\cite{yaghoobi2022prisms}.}\label{fig:ss2}
\end{figure}

\paragraph{Mechanical Response}\leavevmode\\*
Figure~\ref{fig:ss2} overlays the simulated equivalent (von Mises) stress on the PRISMS-Plasticity TM reference curve of Yaghoobi et al.~\cite{yaghoobi2022prisms}. Because loading is imposed through a deviatoric velocity gradient in the Taylor model, the von Mises equivalent stress is the appropriate homogenized flow measure; it is plotted against the true
(logarithmic) strain. The reference curve spans true strain from 0 to 1.0. The simulated curve tracks it closely, with an RMSE of approximately 4.8 MPa and a MAPE of 1.8\%, reaching 384 MPa at a true strain of 1.0 against the reference value of 383 MPa. This close agreement is obtained from a single forward pass with fixed slip parameters. The slight deviations are attributed to the difference between the initial orientation set used in the reference and the one generated by the agent.

\begin{figure}[H]
  \centering
  \includegraphics[width=\linewidth]{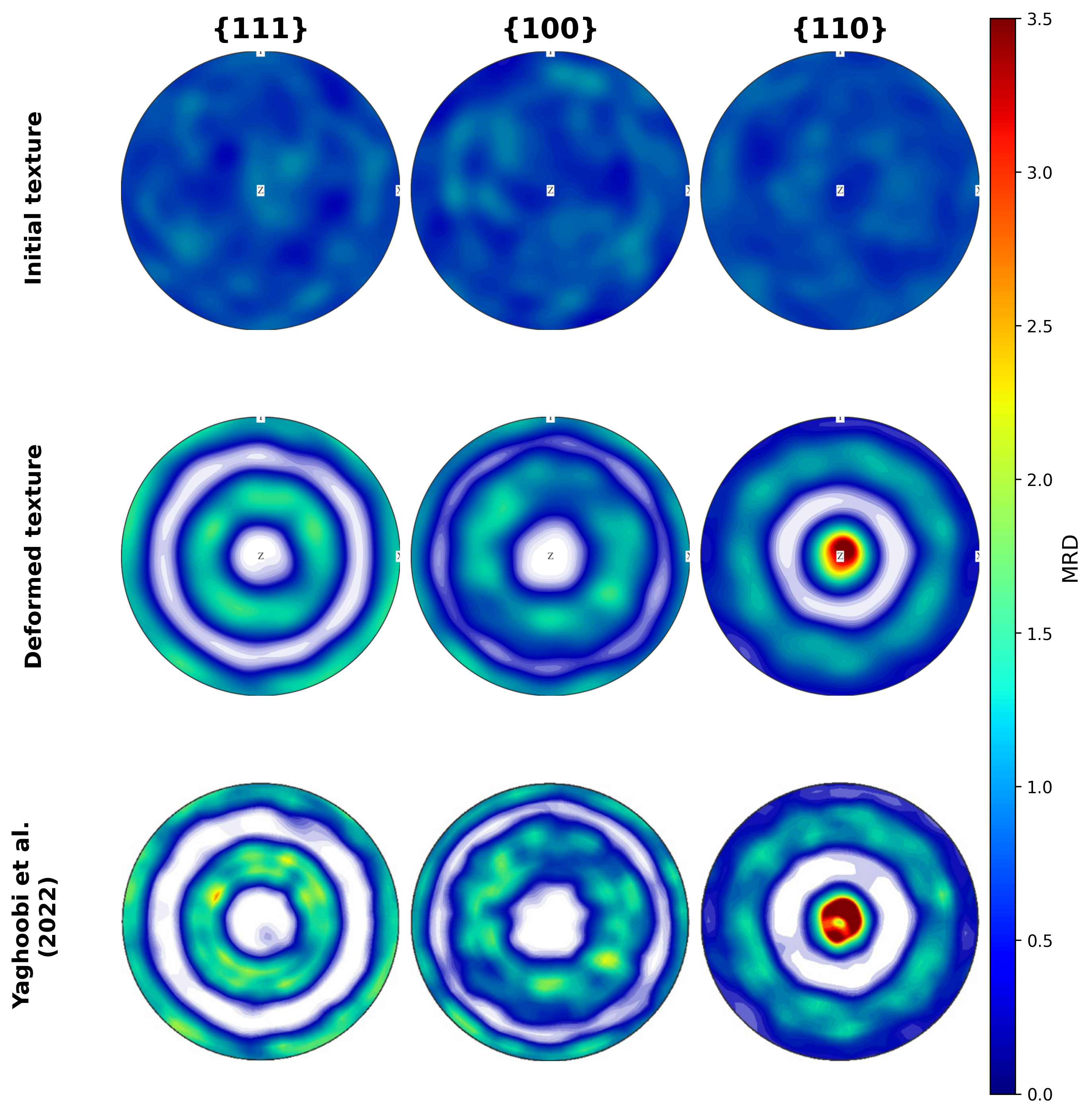}
  \caption{Post-deformation \{111\}, \{100\}, and \{110\} pole figures after 100\% compression on a common intensity scale (MRD 0 to 3.5): this study (top) and Fig.~10 of Yaghoobi et al.~\cite{yaghoobi2022prisms} (bottom), reproduced with permission. The compression axis (Z) is at the center of each projection.}\label{fig:pf2}
\end{figure}

\paragraph{Texture Evolution}\leavevmode\\*
The pre- and post-deformation pole figures (Fig.~S3 in the Supplementary material) reveal a clear deformation texture after compression to a true strain of about 1.0. Prior to deformation, all three pole figures are nearly uniform at approximately one multiples of a random distribution (MRD), confirming the random initial texture. 
Figure~\ref{fig:pf2} compares the post-deformation texture from this study with the corresponding reference results of Yaghoobi et al.~\cite{yaghoobi2022prisms} on a common intensity scale (0 to 3.5 MRD); the compression axis ($Z$) is at the center of each projection.

Both simulations develop the same characteristic deformation pattern: a central $\{110\}$ maximum along the compression axis, the $\langle 110 \rangle$ compression fiber, with central minima in the $\{100\}$ and $\{111\}$ pole figures. 
The agreement is strong at both the pattern and the intensity level. 
In this study, 83\% of the grains lie within $15^\circ$ of the $\langle 110 \rangle \parallel Z$ fiber, and the central $\{110\}$ intensity reaches 3.5 MRD, matching the reference. 
The visible differences lie in local spottiness and symmetry rather than in fiber strength. 
These differences warrant comment because both simulations use comparable grain counts (400 grains each), so grain count cannot explain them. 
Two factors are more plausible. 
First, each texture is a single statistical realization of an approximately 400-grain random aggregate, and the two rely on different initial orientation sets; a 400-grain sample does not populate every azimuth evenly, so its deformed fiber is not perfectly axisymmetric and carries discrete concentrations that differ across realizations. 
Second, and more influential for visual smoothness, is the pole-figure reconstruction: the ODF kernel half-width used in MTEX controls whether discrete grain orientations are blended into a smooth fiber or preserved as individual concentrations, so the same underlying texture can appear markedly smoother or spottier depending on the setting. 
The two studies also used different microstructure-generation packages. 
Since the reference MTEX settings are not published, the observed differences cannot be attributed to a failure of the forward model.

Table~\ref{tab:pole_features} compares the simulated pole-figure features against values read from the reference pole figures of Yaghoobi et al.~\cite{yaghoobi2022prisms}. 
The peak intensities for $\{100\}$, $\{110\}$, and $\{111\}$ are 2.16, 3.75, and 2.23 MRD, respectively. 
The central $\{110\}$ intensity, which measures the $\langle 110 \rangle$ compression-fiber strength, is 3.47 MRD, matching the reference maximum of about 3.5 MRD, while the central $\{100\}$ and $\{111\}$ intensities are close to zero, reproducing the axial depletion seen in the reference.

\begin{table}
\centering
\caption{Quantitative pole-figure features of the post-deformation texture, compared with values read from the reference pole figures of Yaghoobi et al.~\cite{yaghoobi2022prisms} (0 to 3.5 MRD colorbar; reference values approximate, $\pm\sim0.3$ MRD).}
\label{tab:pole_features}
\begin{tabular}{lcc}
\toprule
\textbf{Feature (MRD)} & \textbf{This study} & \textbf{Reference} \\
\midrule
Peak intensity, $\{100\}$ & 2.16 & $\sim$2.0 \\
Peak intensity, $\{110\}$ & 3.75 & $\sim$3.5 \\
Peak intensity, $\{111\}$ & 2.23 & $\sim$2.0 \\
Central $\{110\}$ ($\langle 110 \rangle$ compression fiber) & 3.47 & $\sim$3.5 \\
Central $\{100\}$ (axial depletion) & 0.01 & $\sim$0 \\
\bottomrule
\end{tabular}
\end{table}

\subsubsection{Discussion}\label{sec:cs2disc}
This case study met its objectives without any tuning. 
The macroscopic flow response matched the reference to within the few-percent margin expected of a full-field crystal-plasticity model (MAPE 1.8\%). 
The deformation texture reproduced the correct pattern, the $\langle 110 \rangle$ compression fiber with accompanying $\{100\}$ and $\{111\}$ minima, and, on a matched intensity scale, its peak and central intensities are comparable to the reference (Table~\ref{tab:pole_features}), including a fiber strength (central $\{110\} \approx 3.5$ MRD) that matches rather than underconverges the reference. 
Both results are obtained from a single forward run with fixed parameters, and the agent recovered the full workflow from the goal alone, which is the central capability this framework is intended to demonstrate.

A substep-convergence check ruled out temporal discretization as a source of error: increasing the number of Taylor substeps from 100 to 1000 left both the stress--strain curve and the deformation texture unchanged (identical flow stress and fiber fraction), so the coarser setting is fully converged. 
Because the Taylor model assigns one material point per grain, there is no finite-element mesh to refine, and grain count is comparable between the two studies (400 grains each); neither can therefore explain the residual texture differences of Figure~\ref{fig:pf2}. 
Instead, these arise from the combination of sampling variability and pole-figure reconstruction described above. 
The simulated figures are brought closest to the reference not by finer discretization but by using a broader reconstruction kernel or by ensemble-averaging over several random orientation sets.

The value of this case study is that it anchors the forward model before its use in the inverse setting. 
Having shown that the model reproduces both the flow response and the deformation texture of an independent published benchmark, the same pipeline can be trusted as the forward evaluator in an inverse problem, in which the agent searches the initial microstructural texture to reproduce a target deformation texture. 
The comparison above also sets expectations for the next case study: because the deformed texture of an approximately 400-grain aggregate carries sampling variability and depends on pole-figure reconstruction, the recovered initial texture should be interpreted at the level the forward model supports, not as a precise orientation-by-orientation match. 
This is exactly how the subsequent inverse study interprets its results.

\subsection{Case Study 3: Inverse Recovery of the Initial Texture}\label{sec:cs3}
 Case Study 3 poses a harder inverse problem. Given a target deformation texture, specifically the pole figures of copper after compression to a true strain of approximately 1.0 reported by Yaghoobi et al.~\cite{yaghoobi2022prisms}, the agent is tasked with recovering the initial crystallographic texture that produces it. Whereas Case Study 1 searched a few scalar parameters against a stress-strain curve, this case study searches the entire initial microstructural texture against a full pole-figure target, using the forward model validated in Case Study 2.

\subsubsection{Problem Formulation}\label{sec:cs3form}
The agent is not told the initial texture. The agent searches the initial-texture design variables (Table~\ref{tbl:inv}) so that, after the same uniaxial compression along $Z$ to a true strain of approximately 1.0, the simulated deformed texture matches the target.
The design space is the texture mode, either a random (isotropic) orientation distribution or a fiber aligned to one of the low-index axes $[100]$, $[110]$, $[111]$, and the fiber spread $\sigma$. 
Random is an explicit selectable mode, so a fully isotropic initial texture is directly reachable; increasing the fiber spread makes a fiber progressively more diffuse but does not make it random, because a broad Gaussian spread in the tilt angle retains a weak concentration along the fiber axis. 
Everything else is fixed at the Case Study 2 values: the rate-dependent Taylor copper parameters ($s_0 = 16$ MPa, $h_0 = 200$ MPa, $s_s = 129.5$ MPa, $n = 2$, latent hardening $q = 1.4$, rate sensitivity $m = 77$), the velocity-gradient boundary condition driving the compression, and 400 equal-size grains represented one grain per element on a $5 \times 8 \times 10$ grid (the grain count of the reference, held fixed).

\begin{table}[!ht]
\centering
\caption{Inverse-problem design variables and search configuration.}
\label{tbl:inv}
\begin{tabular*}{\linewidth}{@{\extracolsep{\fill}}p{0.25\linewidth}p{0.65\linewidth}@{}}
\toprule
\textbf{Item} & \textbf{Specification} \\
\midrule
Design variable 1 & Texture mode: random, or fiber about $[100]$, $[110]$, or $[111]$ \\
Design variable 2 & Fiber spread $\sigma$, 3 to 85 degrees (larger spread gives a more diffuse fiber) \\
Objective & Texture-match loss vs target pole figures \\
Search method & Bayesian optimization (Gaussian process + expected improvement) \\
Budget & 15 forward evaluations \\
Fixed & Rate-dependent Taylor Cu parameters ($m = 77$), compression to true strain $\sim 1.0$, 400-grain equal-size RVE \\
\bottomrule
\end{tabular*}
\end{table}

It is computed directly from the deformed orientations, without invoking MTEX for pole-figure evaluation, so that it is fast enough to sit inside the optimization loop. 
For each of the $\{111\}$, $\{100\}$ and $\{110\}$ pole figures the deformed orientations are converted to a pole density in  MRD, smoothed with an 8-degree kernel and normalized so that the mean density is 1, and then reduced to two numbers: the peak intensity, the maximum MRD anywhere in the figure, and the central intensity, the mean MRD within a 12-degree cap around the compression axis $Z$. 
These two numbers are compared with the digitized target values, namely peak intensities of 2.5, 2.35 and 3.5 MRD for $\{111\}$, $\{100\}$ and $\{110\}$ and central intensities of 0.3, 0.3 and 3.3; the $\{110\}$ central value is a maximum, the signature of the compression fiber, while the $\{111\}$ and $\{100\}$ central values are minima. 
The loss is the weighted sum, over the three pole figures, of the squared differences from these targets,
\begin{equation}
L = \sum_{\{hkl\}} w \left( 0.5 \left(\text{peak} - \text{peak}_{\text{target}}\right)^2 + \left(\text{center} - \text{center}_{\text{target}}\right)^2 \right),
\end{equation}
where the weight $w$ is 2 for the $\{110\}$ figure and 1 for the others so that the central $\{110\}$ compression fiber, the most discriminating feature of the target, dominates the objective. 
Each evaluation runs the full forward pipeline (generate the candidate microstructure, convert, compress with the rate-dependent Taylor model, extract the post-deformation orientations) and returns this loss.

\subsubsection{Task Specification}\label{sec:cs3agent}
The agent receives the following user query:

\begin{tcolorbox}[colback=blue!10,colframe=gray!40,boxrule=0.4pt]
\itshape
``Recover the initial crystallographic texture of the copper polycrystal that, after uniaxial compression along $Z$ to true strain $\sim 1.0$, reproduces the target deformation texture in \texttt{fig2c\_targets.json} (the $\{111\}$, $\{100\}$, $\{110\}$ pole figures with a central $\{110\}$ compression fiber). Select the most sample-efficient search strategy and use a budget of 15 simulations. Then interpret the recovered initial texture.''
\end{tcolorbox}

Here, \texttt{fig2c\_targets.json} holds the digitized $\{111\}$, $\{100\}$, and $\{110\}$ pole-figure intensities read from Fig.~2c of the reference, which serve as the target for the inverse search. Any input not specified in the query is treated as a fixed input. The elastic constants and other input parameters are fixed in the PRISMS-Plasticity input file, and the microstructure details for random microstructure generation are fixed in MicroGen. More details on the setup can be found here: \url{https://github.com/samoalfred/harness-cp-agents/tree/main/case_study_1_calibration}.

As in Case Study 1, the agent selects a search strategy from a small repository rather than performing the search itself. 
Offered a Bayesian optimizer and a random-search baseline, it selects Bayesian optimization as the most sample-efficient choice for the expensive forward model and launches it with the requested budget. 
The optimizer then owns the entire search: it proposes candidate initial textures, runs the forward pipeline for each, scores the deformed texture, and updates its surrogate model. 
The agent does not propose textures. 
The first four evaluations form an initial design (random plus each fiber at an intermediate spread); the remaining eleven are chosen by expected improvement.

\subsubsection{Results}\label{sec:cs3res}

\begin{figure}[H]
  \centering
  \includegraphics[width=\linewidth]{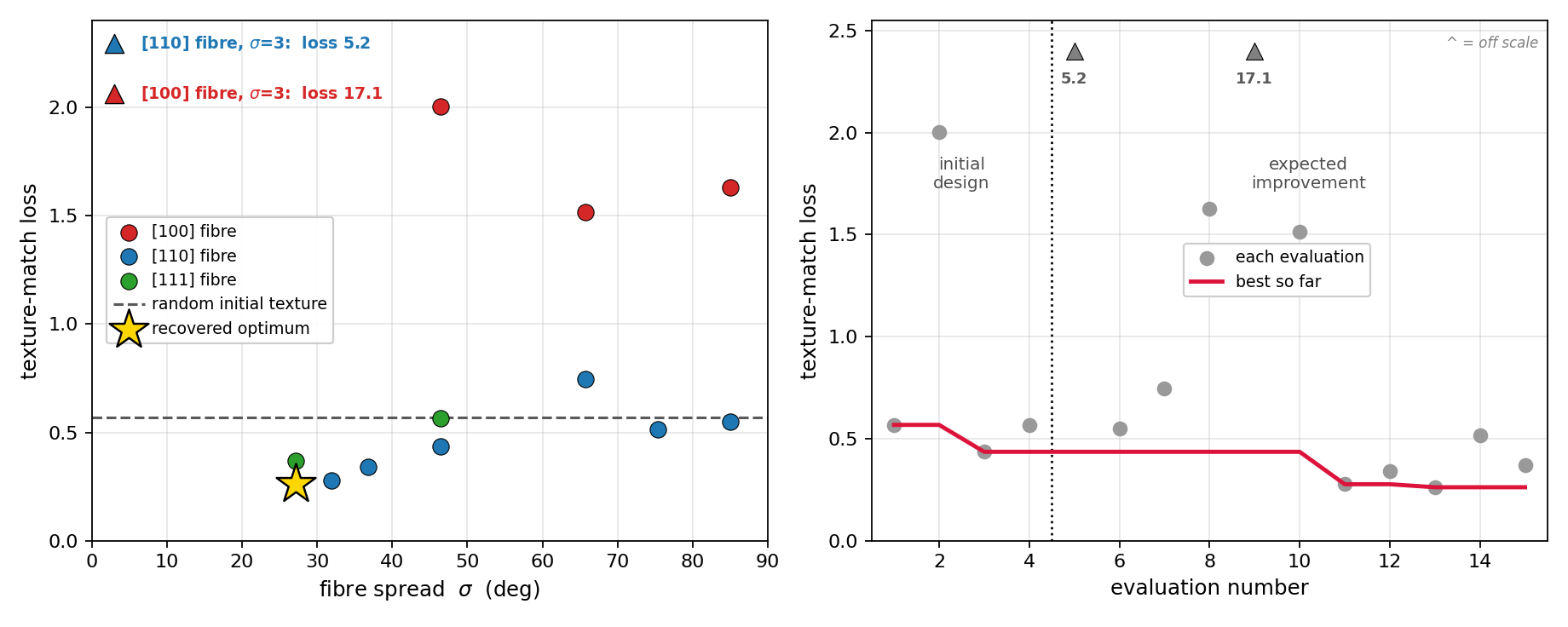}
  \caption{\textit{Outcome of the 15-evaluation inverse search}. (a) Texture-match loss of every evaluated initial texture versus its fiber spread $\sigma$, colored by fiber axis ($[100]$ red, $[110]$ blue, $[111]$ green); the dashed horizontal line is the loss of a random initial texture, the star marks the recovered optimum, and the two upward triangles at the top-left are the sharp fibers ($\sigma = 3$ degrees) whose losses lie above the vertical scale and are printed beside them. (b) The best loss found so far as the search proceeds (red line), with each evaluation shown as a grey point; the dotted vertical line separates the first four evaluations, a fixed initial design (random plus one fiber of each axis), from the remaining eleven, which are chosen by the Gaussian-process expected-improvement acquisition; the two evaluations whose losses exceed the axis are drawn as triangles at the top with their values labeled.}
  \label{fig:search}
\end{figure}

The search outcome is summarized in Fig.~\ref{fig:search} and Table~\ref{tbl:eval}. Figure~\ref{fig:search} is read in two parts. The left sub-figure is a map of the search space: each point is one candidate initial texture; its horizontal position is how diffuse that texture is, from a tight, sharply aligned fiber at small spread to a broad, near-random fiber at large spread; its vertical position is how well the deformed texture it produces matches the target, so lower is better; and its color marks the texture type, black for the random (isotropic) mode and red, blue or green for a fiber aligned to the [100], [110] or [111] axis respectively. Points below the dashed line beat a plain random start, and the star is the best texture found. The right panel tracks the search itself: each point is one simulation in the order it was run, and the solid line is the best score achieved up to that point, which can only fall. The first four points, to the left of the dotted divider, are a fixed spread of starting guesses that survey the space; thereafter the optimizer fits a Gaussian-process surrogate to the results seen so far and places each new evaluation where an improvement is most likely.

As observed from Fig.~\ref{fig:search} and Table~\ref{tbl:eval}, three features are robust. First, sharp initial textures are decisively rejected: a tightly aligned $[110]$ fiber (spread 3 degrees) produces a grossly over-sharp deformed texture with a loss of 5.19, and a sharp $[100]$ fiber a loss of 17.1, one to two orders of magnitude worse than the diffuse candidates.
Second, a $[100]$ initial fiber is rejected at every spread, with losses between 1.5 and 17.
Third, the low-loss region is populated entirely by diffuse textures: fibers about $[110]$ with intermediate to large spreads, a diffuse $[111]$ fiber, and the random mode. The recovered optimum is a moderately diffuse $[110]$ fiber at a spread of 27 degrees, with a loss of 0.262; the converged re-run at full Taylor sub-steps gives the same value (0.262), confirming that the reduced-substep search is accurate. The random mode has a loss of 0.568.
 
 \begin{table}[!ht]
\centering
\caption{\textit{Selected evaluations from the 15-simulation search, ordered by loss}.}
\label{tbl:eval}
\begin{tabular*}{\linewidth}{@{\extracolsep{\fill}}p{0.24\linewidth}p{0.16\linewidth}p{0.12\linewidth}p{0.36\linewidth}@{}}
\toprule
Initial texture & Spread (deg) & Loss & Note \\
\midrule
{}[110] fiber & 27 & 0.262 & Recovered optimum (diffuse) \\
{}[110] fiber & 32 & 0.271 & Diffuse [110], low loss \\
{}[110] fiber & 37 & 0.343 & Diffuse [110], low loss \\
{}[111] fiber & 27 & 0.369 & Diffuse [111], low loss \\
{}[110] fiber & 46 & 0.436 & Diffuse [110] \\
{}[110] fiber & 85 & 0.551 & Very diffuse [110] \\
Random & -- & 0.568 & Near-isotropic; in the low-loss region \\
{}[110] fiber & 66 & 0.747 & Diffuse [110] \\
{}[100] fiber & 66 & 1.516 & Wrong axis \\
{}[100] fiber & 46 & 2.005 & Wrong axis \\
{}[110] fiber & 3 & 5.188 & Sharp fiber, strongly rejected \\
{}[100] fiber & 3 & 17.06 & Sharp, wrong axis, strongly rejected \\
\bottomrule
\end{tabular*}
\end{table}

\begin{figure}[H]
  \centering
  \includegraphics[width=\linewidth]{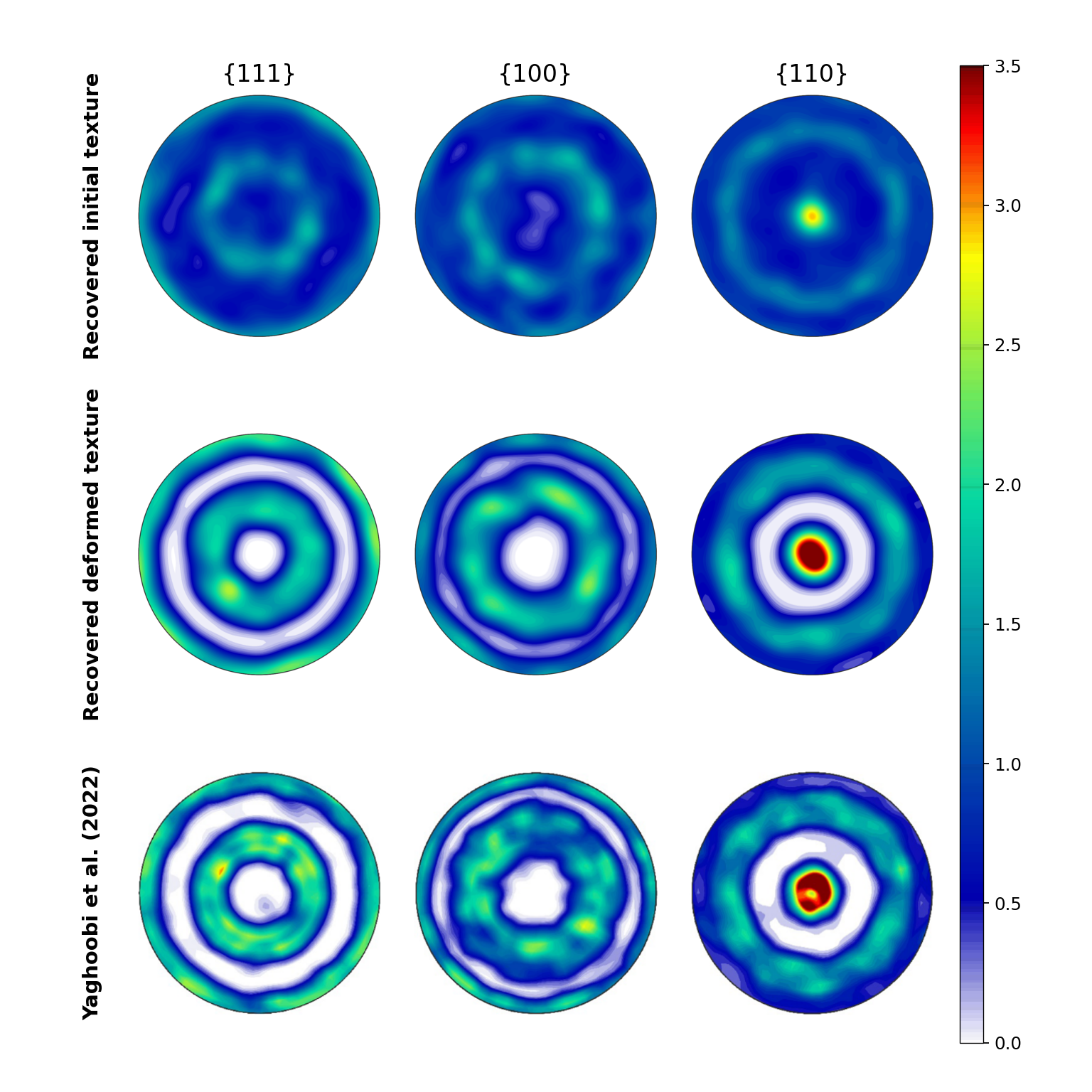}
  \caption{$\{111\}$, $\{100\}$, and $\{110\}$ pole figures on a common intensity scale (MRD 0 to 3.5), top to bottom: the recovered initial texture of this study (a diffuse $[110]$ fiber of spread 27 degrees); its deformed texture after compression to true strain $\sim 1.0$; and the reference deformation texture from Yaghoobi et al. (2022), reproduced with permission. The compression axis ($Z$) is at the center of each projection.}
  \label{fig:pf3}
\end{figure}

Figure~\ref{fig:pf3} places the recovered initial texture, its deformed texture, and the reference on a common intensity scale, top to bottom. 
The top row is the recovered initial texture, the diffuse $[110]$ fiber of spread 27 degrees; the middle row is that same texture after compression to true strain $\sim 1.0$ in this study; and the bottom row is the reference deformation texture from the PRISMS-Plasticity TM study \cite{yaghoobi2022prisms}. 
The weakly $[110]$-biased initial texture develops, under compression, into the target pattern: a central $\{110\}$ maximum along the compression axis, the $\langle 110 \rangle$ compression fiber, with central minima in the $\{100\}$ and $\{111\}$ figures, matching the reference.

The recovered optimum reproduces the pattern and closely approaches the target fiber: its $\{110\}$ central intensity is 3.11 MRD against the 3.3 target, and the $\{111\}$ and $\{100\}$ peaks and central minima bracket the reference.

 \begin{table}[!ht]
\centering
\caption{Quantitative pole-figure features of the recovered optimum's deformed texture. Both columns are evaluated with the pole-figure scorer used in the inverse objective, so the simulated features sit on the same footing as the digitized reference targets~\cite{yaghoobi2022prisms} (values approximate, about $\pm$0.2 MRD).}
\label{tbl:pf3}
\begin{tabular*}{\linewidth}{@{\extracolsep{\fill}}p{0.56\linewidth}p{0.18\linewidth}p{0.16\linewidth}@{}}
\toprule
Feature (MRD) & This study & Reference \\
\midrule
Peak intensity, $\{111\}$ & 2.27 & 2.5 \\
Peak intensity, $\{100\}$ & 2.18 & 2.35 \\
Peak intensity, $\{110\}$ & 3.73 & 3.5 \\
Central $\{110\}$ ($\langle110\rangle$ compression fiber) & 3.11 & 3.3 \\
Central $\{100\}$ (axial depletion) & 0.04 & $\sim$0.3 (min) \\
Central $\{111\}$ (axial depletion) & 0.12 & $\sim$0.3 (min) \\
\bottomrule
\end{tabular*}
\end{table}

\subsubsection{Discussion}\label{sec:cs3disc}
The reconstruction-consistent result is regarded as the principled outcome: when the target central intensities are read with the same MTEX kernel used by the forward scorer, the random mode becomes the global optimum and the [110] preference vanishes, recovering the known ground truth. The [110]
 fiber returned under the  target is best understood as a proxy that compensates the footing gap between the published figure and the simulation's own reconstruction, not as evidence of a physical [110] initial texture.\\
The robust and physically correct result of the search is that the initial texture must be diffuse. 
Every sharp or strongly biased initial texture is rejected: a tight fiber over-sharpens under compression, and a $[100]$ fiber evolves into the wrong pattern. 
The low-loss region is populated by diffuse initial textures, both the explicit random mode and widely spread fibers about $[110]$ and $[111]$. 
This is consistent with the origin of the reference texture, whose initial state is a random orientation distribution.

The recovered optimum, however, is not pure random but a moderately diffuse $[110]$ fiber (spread 27 degrees), and random ranks eighth of the fifteen evaluations under the digitized target. 
This ordering is a property of the measurement footing, not of the physics. 
The objective rewards a strong central $\{110\}$ intensity (digitized target 3.3 MRD), but the forward model produces a central intensity of about 2.9 MRD from a random start when read by the same Python scorer, below the digitized target, for the realization and reconstruction reasons set out in Case Study 2 (a random 400-grain realization and an 8-degree reconstruction kernel), not because of grain count. 
A mild to moderate $[110]$ pre-bias in the initial texture raises the central $\{110\}$ intensity toward the digitized target, so the optimizer prefers it.

To test this directly, the same fifteen evaluations were re-scored against a self-consistent target, in which the central intensities are set to the values the scorer itself assigns to the random ground-truth realization rather than to the digitized figure. 
Under this consistent footing the ordering flips: the random initial texture becomes the global optimum and the preference for a $[110]$ fiber disappears, while sharp and $[100]$ textures remain rejected. 
The recovered $[110]$ component under the digitized target is therefore best read as a proxy that compensates the footing gap, not as evidence of a physical $[110]$ initial texture; the reference initial state is random. 
Notably, the rate-dependent Taylor model reproduces the same qualitative behavior as the earlier rate-independent treatment, which confirms that the residual $[110]$ preference is driven by the reconstruction footing rather than by the choice of forward model.

Read at the level the forward model supports, the inverse search recovers the correct qualitative answer: a diffuse initial texture, with sharp and mis-aligned textures decisively excluded. 
Overall the case study demonstrates the agent solving an inverse problem posed over microstructural texture with a validated rate-dependent forward model: from a natural language goal it selects the search strategy, launches the optimizer, and recovers a diffuse initial texture consistent with the random ground truth, up to a weak $[110]$ bias induced by the measurement footing.

\subsubsection{Limitations}
\label{sec:cs3lim}

The search used fifteen evaluations, so the modest separation between the best diffuse $[110]$ fiber (0.262) and the random mode (0.568) is partly numerical noise from a single 400-grain realization and should not be over-interpreted. 
The objective's $\{110\}$ central target, digitized from the reference figure and read on a footing the model does not match from a random start, biases the search toward pre-biased initial textures; as the re-scoring shows, a target reconstructed on the same kernel as the simulation removes this bias. 
The recovered spread and axis are best reported qualitatively, as a diffuse initial texture, rather than as a precise physical initial state. 
A target calibrated to the simulation's own reconstruction, together with ensemble-averaging over several random realizations and a larger evaluation budget, would be the natural next step to tighten the recovered optimum onto the random ground truth.

\subsection{Case Study 4: Multi-Pass Deformation Texture Evolution in a Magnesium Alloy}\label{sec:cs4}

The first three case studies exercised the harness on single-shot workflows: a bounded parameter search (CS1), one forward pass (CS2), and a search over initial textures evaluated by a single forward pass each (CS3). All three treated the CP simulation as an standalone, stateless operation. The fourth case study removes that assumption. It tasks the agent with a chained, multi-pass workflow in which the output of one simulation becomes the input of the next, and it does so in a different and more demanding material class, a hexagonal-close-packed (HCP) magnesium alloy whose deformation is carried by competing slip families and deformation twinning. The target is the deformation-texture evolution of the dilute alloy ZX31 (Mg--3Zn--0.3Ca) under multi-pass plane-strain compression at $350\,^\circ$C, validated against the PRISMS-Plasticity TM results and rolling experiments of Yaghoobi et al.~\cite{yaghoobi2025effects}. A further distinction from the earlier case studies is the nature of the validation target. CS2 and CS3 checked the forward and inverse texture problems against reference simulation pole figures, whereas CS4 is validated against experimentally measured texture, specifically the $(0001)$ pole figures obtained by electron backscatter diffraction (EBSD) on the physically rolled alloy. The harness is therefore confronted, for the first time in this series, with real characterisation data rather than a simulation benchmark or a synthetic microstructure. As in CS2, no parameters are tuned: the calibrated slip and extension-twinning parameters at $350\,^\circ$C from the reference are fixed inputs, making the run a forward reproduction. What is new is that the agent must recognize the workflow as a stateful sequence of five passes and carry the deformed microstructure from each pass into the next.

\subsubsection{Task specification}\label{sec:cs4task}
The agent receives the following task:

\begin{tcolorbox}[colback=blue!10,colframe=gray!40,boxrule=0.4pt]
\itshape
``Reproduce the multi-pass deformation-texture evolution of ZX31 (Mg-3Zn-0.3Ca) from Yaghoobi, Berman \& Allison (2025) with PRISMS-Plasticity. This is a chained, five-pass workflow simulating hot rolling: each plane-strain-compression pass deforms the material to 20\% true strain along the normal direction, and the deformed grain orientations from one pass become the initial texture of the next pass. The alloy is HCP with the calibrated slip/twin parameters already set in prm.prm, and it requires the RATE-DEPENDENT constitutive model. Run the five passes in order. After the FIRST
pass, compare the single-pass mechanical response against the reference simulation, reporting RMSE and MAPE. After the FIFTH pass, analyse the (0001) basal texture evolution (peak intensity vs pass number) and compare the final texture against the reference 5-pass simulation and against the experiment. Finally, interpret the result.''
\end{tcolorbox}

The elastic constants and the rate-dependent slip parameters for basal, prismatic and pyramidal $\langle c+a \rangle$ families are omitted from the query because they are fixed inputs in the PRISMS-Plasticity file. More details on the setup can be found here: \url{https://github.com/samoalfred/harness-cp-agents/tree/main/case_study_4_multipass}

\subsubsection{Harness configuration}\label{sec:cs4config}
As in the other case studies, CS4 uses the goal-driven ReAct configuration: the agent is given the objective and the tool schemas only, and must infer the operation sequence itself. The tool layer exposes four typed operations. The central one, \texttt{run\_pass(pass\_number)}, is a stateful wrapper around a single rate-dependent PRISMS-Plasticity pass. For pass one it reads the random as-cast orientations; for pass $N$ it reads the deformed orientations produced by pass $N-1$. On completion it writes that pass's per-quadrature-point output, records the plane-strain flow stress and the reoriented twin fraction, and for every pass except the last, reformats the deformed lattice orientations (Rodrigues components, aggregated by grain) into the orientation-input file consumed by the next pass. This output-to-input hand-off is the state carried across the chain, and it is performed deterministically inside the tool rather than by the LLM. The remaining three tools are read-only analyses: \texttt{compare\_stress\_strain} (pass-1 flow stress versus the reference, RMSE and MAPE), \texttt{analyze\_texture\_evolution} (the $(0001)$ peak intensity as a function of pass number), and \texttt{compare\_final\_texture} (the pass-five pole figure against the reference simulation and the experiment on a matched intensity scale).

The design principle is unchanged from CS1--CS3: the LLM is responsible for recognizing the workflow and sequencing the operations, while the numerically and physically consequential steps, the finite-strain integration and, critically, the pass-to-pass orientation hand-off, are delegated to validated, deterministic tools. The agent must reason that the five passes are ordered and mutually dependent (pass $N$ cannot run until pass $N-1$ has produced its deformed texture), that the single-pass stress comparison is due after the first pass, and that the texture-evolution analyses are due only once all five passes are complete. This ordering is not prescribed in the prompt; it is inferred from the prerequisites declared in each tool schema, exactly as in CS2, but now over a longer, stateful chain.

\subsubsection{Results}\label{sec:cs4res}
From the goal and the tool descriptions alone, the agent recovered the correct chained workflow: it ran the five passes in order, invoked the single-pass stress comparison after pass one, and, after pass five, produced the texture-evolution analysis and the final-texture comparison. 

\paragraph{Mechanical Response}\leavevmode\\*
The pass-1 plane-strain flow stress reproduced the reference simulation closely ($138.5$ MPa at 20\% true strain, against 138.8 MPa for the reference), and the flow stress rose monotonically from pass to pass as the aggregate work-hardened across the chain. Because the mechanical agreement mirrors the CS2 result and is not the object of this case study, the full pass-1 stress--strain curve and its error metrics are relegated to Section~S4 of the Supplementary Material. 

\begin{figure}[H]
  \centering
  \includegraphics[width=0.7\linewidth]{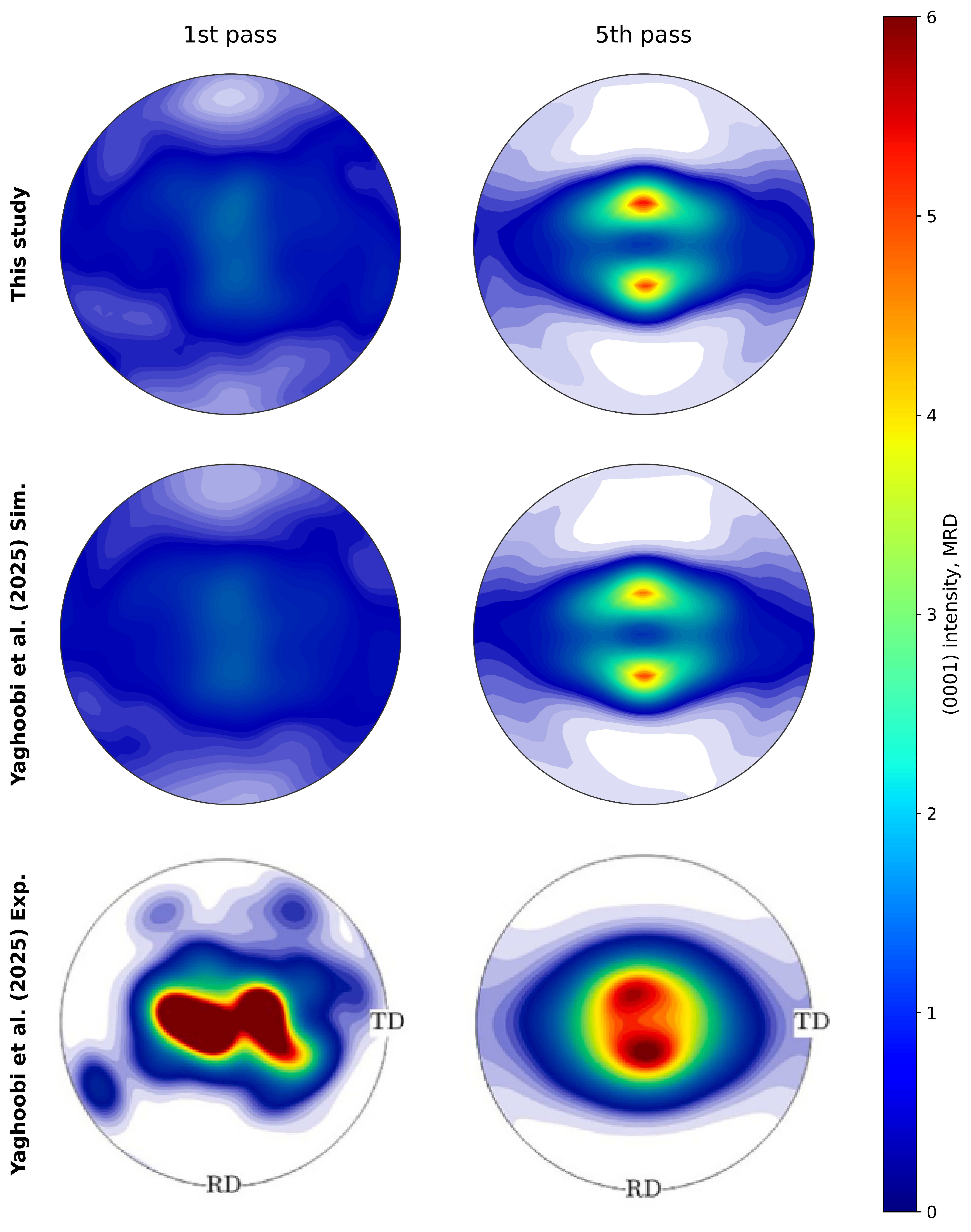}
  \caption{\textit{Evolution of the ZX31 $(0001)$ basal texture under multi-pass plane-strain compression at $350\,^\circ$C}. Columns: after the first pass (left) and the fifth pass (right). Rows: this study (top), reference simulation of Yaghoobi et al.~\cite{yaghoobi2025effects} (middle), and rolling experiment (bottom). The normal direction is at the centre of each projection; RD is vertical (bottom) and TD horizontal (right).}\label{fig:cs4tex}
\end{figure}

It is worth clarifying why the mechanical validation is made against a simulation rather than against experiment. In the reference study the slip and twinning parameters were calibrated to single-pass plane-strain compression curves measured at a higher strain rate (0.5 s$^{-1}$ at $350\,^\circ$C; Fig.~2 of Yaghoobi et al.~\cite{yaghoobi2025effects}), and those fixed parameters were then applied to the multi-pass rolling modelled here, which proceeds at the lower rate imposed by the Gleeble velocity gradient (0.01 s$^{-1}$). Because the flow rule is rate dependent, the rolling response sits below the higher-rate calibration curve, so the experimental stress--strain data are not rate-matched to the rolling simulation and are not overlaid on it. The mechanical check is therefore made against the authors' rolling simulation at the matched rate, while the experimental comparison in this case study is made at the level of texture (Fig.~\ref{fig:cs4tex}). The two tests share the same deformation mode (plane-strain compression) and the same material, but differ in strain rate (0.5 versus 0.01 s$^{-1}$), in the number of passes (a single calibration pass versus five accumulated rolling passes), and in the measured quantity (a stress--strain curve for calibration versus a deformation texture for rolling). This calibrate-then-apply separation is standard practice: the parameters are fixed by one test and the model is exercised, without further tuning, on a different loading history.

\paragraph{Texture Evolution}\leavevmode\\*
Figure~\ref{fig:cs4tex} is the central result. It compares the $(0001)$ basal pole figure of ZX31 after the first and fifth passes for the present agent-driven simulation (top row), the reference simulation of Yaghoobi et al.~\cite{yaghoobi2025effects} (middle row), and the corresponding rolling experiment (bottom row), with the normal direction at the centre of each projection and a common intensity scale for the two simulation rows. Both simulations develop the same texture character: a weak basal pole that splits away from the centre toward the rolling direction, rather than the single strong basal peak characteristic of conventional magnesium. The present simulation reproduces the reference simulation closely at both passes (Table~\ref{tab:cs4tex}): the $(0001)$ peak intensity is 2.10 versus 1.94 MRD after one pass and 5.66 versus 5.46 MRD after five passes. The texture strengthens monotonically with pass number yet remains weak even after five passes, the signature of a Ca-containing alloy.

\begin{table}
\centering
\caption{$(0001)$ peak intensity (MRD) of ZX31 versus pass number.}
\label{tab:cs4tex}
\begin{tabular}{cccc}
\toprule
\textbf{Pass} & \textbf{This study} & \textbf{Yaghoobi et al.~\cite{yaghoobi2025effects} Sim.} & \textbf{Experiment~\cite{yaghoobi2025effects}} \\
\midrule
1 & 2.10 & 1.94 & 7.3 \\
5 & 5.66 & 5.46 & 4.6 \\
\bottomrule
\end{tabular}
\end{table}
The experimental panels deserve a closer look. After five passes the experiment shows the same weak, RD-split basal character as both simulations, but at a lower peak intensity (4.6 MRD) than the deformation-only simulation (5.66 MRD). This ordering is expected: the simulation models deformation alone, whereas the physical material undergoes dynamic recrystallisation during hot rolling, which consumes deformed orientations and weakens the texture. The larger discrepancy after a single pass (experiment 7.3 MRD against simulation 2.10 MRD) reflects a different origin, namely that the experiment begins from an already-textured wrought material while the simulation begins from a random as-cast aggregate; the two converge in character as the imposed rolling strain accumulates and the deformation texture comes to dominate the starting texture. The agreement is therefore strong at the level the deformation-only forward model supports, namely texture character and the simulation-to-simulation intensities, rather than as an orientation-by-orientation match to the experiment, consistent with the interpretation adopted in CS2 and CS3.

\subsubsection{Discussion}\label{sec:cs4disc}
CS4 extends the framework along three axes at once. First, it moves from single-shot to chained execution: the agent orchestrates a five-pass sequence in which each pass depends on the deformed state produced by its predecessor. The stateful hand-off, carrying the deformed orientations of one pass into the next, is exactly the kind of bookkeeping that is error-prone when done by hand and that the harness performs deterministically inside a single tool, leaving the agent to reason only about sequencing and stopping. Second, it moves from FCC to HCP plasticity, where the response is governed by strongly anisotropic slip families and by deformation twinning; the same harness, with only the alloy-specific deck and tools swapped in, reproduces the reference without modification to the agent or its loop. A third distinction is the validation target: unlike the earlier texture case studies, which were benchmarked against reference simulations, CS4 is measured against experimentally determined EBSD textures of the rolled alloy, so the harness is tested here against real characterization data.

The physical result recovers the alloy-design message of the reference. The high basal-to-pyramidal and twin-to-pyramidal critical-resolved-shear-stress ratios of the Ca-containing alloy suppress basal slip and extension twinning in favour of pyramidal $\langle c+a \rangle$ slip, so ZX31 develops a weak basal texture that splits toward the rolling direction and stays weak with increasing passes, in contrast to the strong single-peak basal texture of conventional magnesium. That the agent-driven pipeline reproduces this behaviour, and reproduces the reference simulation to within the sampling margin expected of a finite-grain aggregate, demonstrates that the harness generalises beyond the FCC, single-pass setting of CS2 to the stateful, multi-pass, twinning-active workflows that dominate practical thermomechanical processing.

\subsection{Harness Reliability and Robustness}
\label{sec:harness_robustness}

\subsubsection{Reliability}
Each case study was executed 20 times under its deployed configuration (GPT-4o, temperature 0.1). The agent recovered the correct tool sequence and terminated cleanly in 20 of 20 runs for each of the three repeated-run case studies (CS1--CS3); run-to-run variation was confined to the ordering of mutually independent steps and never violated a data dependency. Variation in the final calibration metrics reflects optimizer stochasticity rather than agent variability. The code to reproduce this section is available at \url{https://github.com/samoalfred/harness-cp-agents/tree/main/experiments}.

\subsubsection{Sequencing Ablation}
To probe where the workflow ordering originates, we crossed tool-name informativeness (descriptive vs opaque \texttt{tool\_a}\ldots\texttt{tool\_h}) with tool-description content (full prerequisites vs prerequisites removed) in a $2 \times 2$ design, holding the minimal goal-driven system prompt and the task query fixed so that the tool layer is the only variable. 
Twenty independent runs were executed per cell (GPT-4o, temperature 0.1) and scored for whether the emitted tool sequence respected every data dependency. 
The agent produced a valid execution order in 20 of 20 runs in all four conditions, including the opaque-name, prerequisite-free cell. 
The ordering is therefore robust to removing the explicit prerequisite clauses, to hiding the tool names, and to removing both simultaneously, indicating that the order is redundantly encoded across the tool names, the functional semantics of the descriptions, and the task statement rather than depending on any single cue. 
Together with the reliability result above, this shows the harness reliably realizes the intended workflow from a natural-language goal and is not brittle to the phrasing of the tool layer.

\subsubsection{Error Recovery}
The single-agent design retains the full tool history so the agent can diagnose a failure and adapt. 
To demonstrate this, a controlled fault was injected into the CS2 workflow: the solver was configured to abort with a diagnostic reporting that the microstructure had not been converted to PRISMS-Plasticity input. 
In 5 of 5 runs the agent read the failure observation, inferred the missing conversion step, invoked the conversion tool, re-ran the solver successfully, and completed the remaining pipeline. 
Notably, the agent overrode the default ``do not convert'' instruction in its prompt once the solver evidence contradicted it, showing that recovery is driven by reasoning over the observed result rather than by a fixed sequence. The code to reproduce this section is available on GitHub: \url{https://github.com/samoalfred/harness-cp-agents/tree/main/experiments/recovery}.

\section{Discussion}
\label{sec:discussion}

The four case studies collectively demonstrate that CP-Agent, a harness-engineered LLM-based agent can autonomously execute CP modeling workflows spanning forward prediction, parameter calibration, inverse texture recovery, and multi-pass texture evolution. The same harness architecture, with no modification to the system prompt, generalized across these tasks; only the tool set and user query changed. This problem-agnostic capability distinguishes the present framework from prior work that hard-codes workflow steps in prompts or relies on domain-specific fine-tuning.

\subsection{Harness Design and Generalization}

The central design principle of this work is that the harness, not the model, is the primary engineering artifact. The system prompt provides only a goal and behavioral rules; the tool descriptions, together with the user's natural-language task statement, carry the domain knowledge that supports autonomous sequencing. 
The agent inferred the correct workflow in Case Study 2 without any explicit ordering instruction, and an ablation (Section~\ref{sec:harness_robustness}) shows this behavior is robust to the phrasing of the tool layer. This separation of concerns ensures that adapting the framework to new materials, solvers, or post-processing requirements requires only updating the tool layer, not rewriting the prompt or agent loop. The same harness that calibrated slip parameters in Case Study 1 also recovered an initial texture in Case Study 3, demonstrating that the approach is not overfitted to a single problem class.

\subsection{Role of the Agent versus the Optimizer}

A deliberate design choice in this framework is the clear boundary between agent reasoning and numerical search. The agent selects the strategy, sequences the tools, and interprets the results, but it never proposes parameter values or executes the search itself. This division of labor is critical for reliability. Numerical search is a well-understood problem with established algorithms that do not benefit from LLM judgment; inserting the LLM into the optimization loop would introduce stochastic variability and risk of non-physical proposals. By delegating the search entirely to validated optimizers, the harness ensures that the numerical component is both reproducible and physically bounded. The agent's reasoning capacity is spent where it adds value: choosing the appropriate algorithm, diagnosing failures, and providing physical interpretation.

\subsection{Interpretation of the Inverse Texture Result}

The inverse texture case study illustrates both the power and the limitations of the approach. The agent correctly recovered the qualitative result that the initial texture must be diffuse, rejecting sharp fibers and misaligned axes. The recovered optimum, however, was a diffuse [110] fiber rather than pure random, which is the known ground truth. This residual bias is not a failure of the agent but a consequence of the forward model's shortfall relative to the digitized reference, a shortfall attributable to sampling variability and pole-figure reconstruction effects identified in Case Study 2. The result underscores an important principle for inverse problems in this setting: the recovered initial texture should be interpreted at the level the forward model supports, not as a precise orientation-by-orientation match. The agent's final interpretation correctly reflected this caveat, demonstrating that the harness can guide the model toward physically meaningful conclusions even when the numerical objective suggests a slightly biased optimum.

\subsection{Future Directions}

The present work opens several avenues for future development. First, the harness architecture can be extended to multi-scale coupling, where the agent orchestrates simulations across atomistic, mesoscale, and continuum levels, using the output of one scale as the input to the next. Second, the tool layer can be expanded to include uncertainty quantification, enabling the agent to report confidence intervals alongside parameter estimates. Third, the framework can be integrated with experimental data acquisition systems, allowing the agent to autonomously design and execute experiments informed by simulation results. Fourth, the harness can be extended to multi-agent configurations for parallelizable subtasks, such as simultaneous evaluation of multiple candidate microstructures, while retaining the single-agent architecture for sequential decision-making. Fifth, the tool description methodology can be formalized as a reusable knowledge base for CP modeling workflows, reducing the effort required to adapt the framework to new materials or constitutive models. 

\section{Conclusion}
\label{sec:conclusion}

This study presented \textbf{CP-Agent}, a harness-engineered LLM-based agent framework for autonomous CP modeling. The framework combines a minimal \textit{ReAct}-style system prompt with typed tool definitions that encode prerequisites and defaults, a dispatcher, and a safety-bounded iteration loop. The harness is problem-agnostic: the same architecture, with only the tool set and user query changed, successfully executed four distinct workflows without modification to the system prompt or orchestration logic.

Three main conclusions follow. First, harness engineering provides a systematic and generalizable approach to automating CP modeling workflows. The tool descriptions, not the system prompt, carry the domain knowledge that enables autonomous sequencing, ensuring that adaptation to new problems requires only updating the tool layer. Second, the clear division of labor between agent reasoning and numerical search is essential for reliability. The agent selects strategies and interprets results, while validated optimizers execute the search, preventing the LLM from introducing stochastic variability into numerical optimization. Third, the visible reasoning trace required by the \textit{ReAct} paradigm provides an auditable record of each decision, addressing the opacity limitation identified in prior work and establishing trust in autonomous scientific workflows.

The use of CP-Agent was demonstrated on four case studies. In the parameter calibration case, the agent identified four slip parameters for stainless steel 316L, achieving RMSE below 5 MPa within 26 of a 60-simulation budget. In the forward validation case, the agent executed the complete eight-step pipeline for copper, reproducing both the stress-strain response and the deformation texture of an independent published benchmark. In the inverse texture case, the agent recovered a diffuse initial texture consistent with the random ground truth, with residual bias correctly attributed to sampling variability and pole-figure reconstruction effects rather than to model failure.In the multi-pass texture-evolution case, the agent chained five plane-strain rolling passes of the magnesium alloy ZX31, reproducing the reference simulation and the experimentally observed weakened, split basal texture.

By demonstrating that a single harness can generalize across forward and inverse problems in CP modeling, this work establishes a foundation for broader adoption of harnessed-engineered agents in multiscale modeling. The framework reduces the manual effort required to configure, execute, and interpret CP simulations, making these powerful methods more accessible to non-specialists while maintaining the physical interpretability and auditability essential for scientific credibility. Future work will extend the harness to multi-scale coupling, uncertainty quantification, and automated experimental design, progressively reducing human intervention in the materials simulation pipeline while preserving the reasoning and judgment that remain essential to scientific discovery.

\section*{Appendix}
\label{sec:Appendix}

\section*{Reasoning Trace:  Case Study 1}

\begin{lstlisting}
samoalfred@SAMOALFRED:~/candi/plasticity/applications/crystalPlasticity/fcc/SS316L_Semi_LLM_React_MicroGen3$ python3.7 calibrate_react.py

=================================================================
  SS316L ReAct Calibration Agent
  Model     : GPT-4o
  Role      : LLM selects optimizer from repository
  Optimizers: Bayesian | DE | Nelder-Mead | Random+BO
  Budget    : 60 simulations | RMSE<5 MPa | MAPE<2%
=================================================================

=================================================================
@R@[ReAct] SS316L REACT CALIBRATION AGENT@r@
@R@[ReAct] Query: Generate a random SS316L microstructure for crystal plasticity simulations using PRISMS plasticity. Then calibrate the crystal plasticity slip parameters for SS316L to match the experimental tensile stress-strain data (SS316L_experiment.txt). Use Bayesian Optimization. Tune: Initial Slip Resistance [100-150 MPa], Initial Hardening Modulus [800-2500 MPa], Saturation Stress [350-600 MPa], Power Law Exponent [1-3]. Stop when RMSE < 5 MPa or MAPE < 2% or after 60 simulations.@r@
=================================================================

@R@[ReAct] --- Iteration 1 ---@r@

@R@[ReAct] Thought: First, I will generate the synthetic 3D SS316L polycrystal microstructure using MATLAB, which is necessary for the calibration simulations. Then, I will convert the generated HDF5 microstructure into the PRISMS-Plasticity input format. After preparing the microstructure, I will proceed with Bayesian Optimization to calibrate the slip parameters. Let's begin by generating the microstructure and converting it.@r@
@R@[ReAct] Action: generate_microstructure({})@r@
[MATLAB] Running microstructure_gen.m ...
[MATLAB] Working dir: \\wsl.localhost\ubuntu-18.04\home\samoalfred\candi\plasticity\applications\crystalPlasticity\fcc\
SS316L_Semi_LLM_React_MicroGen3\matlab
[MATLAB] This may take several minutes for a 300x300x300 grid.
[MATLAB] >>> USER STARTUP EXECUTED <<<
[MATLAB] Warning: Unable to obtain a change notification handle for
[MATLAB] \\wsl.localhost\ubuntu-18.04\home\samoalfred\candi\plasticity\applications\crystalPlasticity\fcc\
SS316L_Semi_LLM_React_MicroGen3\matlab.]
[MATLAB] initialize MTEX 6.0.0  ... done!
[MATLAB] Warning: No pore file found. Proceeding with no pores.]
[MATLAB] > In microstructure_gen (line 81)
[MATLAB] In run (line 112)]
[MATLAB] === PARALLEL + SPATIAL FILTERING MICROSTRUCTURE GENERATION ===
[MATLAB] Starting parallel pool...
[MATLAB] Starting parallel pool (parpool) using the 'Processes' profile ...
[MATLAB] Connected to parallel pool with 8 workers.
[MATLAB] Parallel pool started in 16.7s with 8 workers
[MATLAB] Building spatial acceleration structure...
[MATLAB] Spatial acceleration structure built in 0.4s
[MATLAB] Processing 1 blocks (1x1x1) in parallel:
[MATLAB] Using 1 batches of 1 blocks each
[MATLAB] Starting parallel processing:
[MATLAB] Batch 1/1: 100.0% | 7653.0 grains/block | Elapsed: 17.7s | ETA: 0.0s
[MATLAB] === COMPLETION SUMMARY ===
[MATLAB] Total time: 17.7 seconds (0.29 minutes)
[MATLAB] SUCCESS: All voxels assigned.
[MATLAB] No gas voids.
[MATLAB] Generating Random Texture
[MATLAB] Saved HDF5: input_structure_poly.h5
[MATLAB] Saved params .mat
[MATLAB] WARNING: Full .mat save failed (large file on WSL path): Unable to write to file '\\wsl.localhost\ubuntu-18.04\home\samoalfred\candi\plasticity\applications\crystalPlasticity\fcc\
SS316L_Semi_LLM_React_MicroGen3\matlab\xct_poly_params_SS316L_full.mat' because it appears to be corrupt.
[MATLAB] IPF coloring computed successfully.
[MATLAB] Output summary written.
[MATLAB] Warning: Exported image displays axes toolbar. To remove axes toolbar from image, export again.]
[MATLAB] Pre-deformation pole figures saved.
[MATLAB] All figures saved to: \\wsl.localhost\ubuntu-18.04\home\samoalfred\candi\plasticity\applications\crystalPlasticity\fcc\
SS316L_Semi_LLM_React_MicroGen3\matlab\figures
[MATLAB] Done!
[MATLAB] Total computation time: 55.42 seconds (0.92 minutes)
[MATLAB] Parallel pool using the 'Processes' profile is shutting down.
[MATLAB] Finished in 72.0 seconds.
[MATLAB] Output summary:
Total number of Grains: 7653, Pores: 0
Domain size: 1.0 x 1.0 x 1.0 mm^3
Domain grid: 32 x 32 x 32 = 32768 voxels
Voxel size: 0.03226 mm (32.26 um)
Grain voxels: 32768 (100.000%)
Pore  voxels: 0 (0.000%)
Total voxels: 32768
Mean grain size (XY): 55.06 um
Mean aspect ratio (Z/XY): 1.00
Done!

@R@[ReAct] Observation: SUCCESS: Microstructure generated successfully (72.0s).@r@
@R@[ReAct] Action: convert_hdf5_to_prisms({})@r@
[H5] Reading: /home/samoalfred/candi/plasticity/applications/crystalPlasticity/fcc/SS316L_Semi_LLM_React_MicroGen3
/matlab/input_structure_poly.h5
[H5] Grid: 32x32x32 | Grains: 7653
[H5] Writing grainID.txt ...
[H5] grainID.txt written (1024 rows x 32 cols).
[H5] Writing orientations.txt ...
[H5] orientations.txt written (7653 grains).
[H5] prm.prm updated: Voxels = 32x32x32
@R@[ReAct] Observation: SUCCESS: Conversion complete: 32x32x32 grid, 7653 grains. | Grid: 32x32x32, 7653 grains. grainID.txt and orientations.txt written; prm.prm updated.@r@

@R@[ReAct] --- Iteration 2 ---@r@
@R@[ReAct] Action: run_bayesian_optimization({'n_initial': 10, 'seed': 42})@r@

[Tool] Calling run_bayesian_optimization({'n_initial': 10, 'seed': 42})

[Tool 1] Bayesian Optimization (LHS + GP+EI)
  [LHS] s0=140.78 h0=1909.21 ss=564.08 n=1.594
  [sim] Eval 1: s0=141 h0=1.91e+03 ss=564 n=1.59
  [sim] RMSE=26.018 MPa  MAPE=8.18%
  [LHS] s0=105.29 h0=1207.98 ss=534.64 n=1.155
  [sim] Eval 2: s0=105 h0=1.21e+03 ss=535 n=1.16
  [sim] New best saved: RMSE=19.7160 MPa
  [sim] RMSE=19.716 MPa  MAPE=6.62%
  [LHS] s0=129.33 h0=2167.93 ss=475.40 n=1.988
  [sim] Eval 3: s0=129 h0=2.17e+03 ss=475 n=1.99
  [sim] New best saved: RMSE=13.2500 MPa
  [sim] RMSE=13.250 MPa  MAPE=5.13%
  [LHS] s0=103.01 h0=2155.54 ss=430.77 n=2.779
  [sim] Eval 4: s0=103 h0=2.16e+03 ss=431 n=2.78
  [sim] RMSE=22.534 MPa  MAPE=7.39%
  [LHS] s0=138.54 h0=2369.57 ss=381.03 n=2.320
  [sim] Eval 5: s0=139 h0=2.37e+03 ss=381 n=2.32
  [sim] RMSE=22.749 MPa  MAPE=7.43%
  [LHS] s0=110.10 h0=1665.40 ss=367.08 n=2.184
  [sim] Eval 6: s0=110 h0=1.67e+03 ss=367 n=2.18
  [sim] RMSE=14.225 MPa  MAPE=5.08%
  [LHS] s0=149.85 h0=1075.13 ss=590.25 n=2.818
  [sim] Eval 7: s0=150 h0=1.08e+03 ss=590 n=2.82
  [sim] RMSE=33.999 MPa  MAPE=9.99%
  [LHS] s0=124.16 h0=1375.02 ss=520.83 n=1.639
  [sim] Eval 8: s0=124 h0=1.38e+03 ss=521 n=1.64
  [sim] New best saved: RMSE=7.2725 MPa
  [sim] RMSE=7.273 MPa  MAPE=3.47%
  [LHS] s0=116.06 h0=967.15 ss=404.33 n=2.409
  [sim] Eval 9: s0=116 h0=967 ss=404 n=2.41
  [sim] RMSE=8.651 MPa  MAPE=3.46%
  [LHS] s0=130.91 h0=1559.35 ss=459.78 n=1.265
  [sim] Eval 10: s0=131 h0=1.56e+03 ss=460 n=1.27
  [sim] RMSE=14.713 MPa  MAPE=5.49%
  [GP+EI] s0=118.81 h0=941.95 ss=544.29 n=2.117
  [sim] Eval 11: s0=119 h0=942 ss=544 n=2.12
  [sim] New best saved: RMSE=5.9381 MPa
  [sim] RMSE=5.938 MPa  MAPE=2.97%
  [GP+EI] s0=126.58 h0=855.52 ss=468.90 n=1.815
  [sim] Eval 12: s0=127 h0=856 ss=469 n=1.82
  [sim] RMSE=8.898 MPa  MAPE=3.91%
  [GP+EI] s0=120.92 h0=1408.28 ss=498.45 n=2.151
  [sim] Eval 13: s0=121 h0=1.41e+03 ss=498 n=2.15
  [sim] New best saved: RMSE=5.1669 MPa
  [sim] RMSE=5.167 MPa  MAPE=2.92%
  [GP+EI] s0=110.26 h0=1628.04 ss=569.56 n=2.055
  [sim] Eval 14: s0=110 h0=1.63e+03 ss=570 n=2.05
  [sim] RMSE=12.846 MPa  MAPE=4.79%
  [GP+EI] s0=120.21 h0=1176.55 ss=478.01 n=2.098
  [sim] Eval 15: s0=120 h0=1.18e+03 ss=478 n=2.1
  [sim] RMSE=5.278 MPa  MAPE=2.94%
  [GP+EI] s0=123.29 h0=1205.81 ss=522.46 n=2.131
  [sim] Eval 16: s0=123 h0=1.21e+03 ss=522 n=2.13
  [sim] RMSE=6.325 MPa  MAPE=3.19%
  [GP+EI] s0=117.59 h0=1195.41 ss=504.22 n=2.018
  [sim] Eval 17: s0=118 h0=1.2e+03 ss=504 n=2.02
  [sim] RMSE=6.307 MPa  MAPE=2.98%
  [GP+EI] s0=128.06 h0=1374.18 ss=367.91 n=2.887
  [sim] Eval 18: s0=128 h0=1.37e+03 ss=368 n=2.89
  [sim] RMSE=10.237 MPa  MAPE=4.29%
  [GP+EI] s0=124.89 h0=1527.68 ss=431.51 n=2.291
  [sim] Eval 19: s0=125 h0=1.53e+03 ss=432 n=2.29
  [sim] RMSE=7.649 MPa  MAPE=3.56%
  [GP+EI] s0=122.17 h0=846.17 ss=527.53 n=2.770
  [sim] Eval 20: s0=122 h0=846 ss=528 n=2.77
  [sim] RMSE=5.719 MPa  MAPE=3.06%
  [GP+EI] s0=118.85 h0=947.86 ss=467.41 n=2.504
  [sim] Eval 21: s0=119 h0=948 ss=467 n=2.5
  [sim] RMSE=6.224 MPa  MAPE=3.02%
  [GP+EI] s0=120.11 h0=850.05 ss=538.35 n=2.944
  [sim] Eval 22: s0=120 h0=850 ss=538 n=2.94
  [sim] RMSE=5.775 MPa  MAPE=3.01%
  [GP+EI] s0=122.20 h0=1520.58 ss=485.64 n=2.964
  [sim] Eval 23: s0=122 h0=1.52e+03 ss=486 n=2.96
  [sim] RMSE=5.659 MPa  MAPE=3.05%
  [GP+EI] s0=123.86 h0=807.27 ss=593.93 n=1.699
  [sim] Eval 24: s0=124 h0=807 ss=594 n=1.7
  [sim] RMSE=6.616 MPa  MAPE=3.26%
  [GP+EI] s0=118.25 h0=847.47 ss=593.52 n=2.639
  [sim] Eval 25: s0=118 h0=847 ss=594 n=2.64
  [sim] RMSE=6.569 MPa  MAPE=3.05%
  [GP+EI] s0=120.61 h0=2398.67 ss=541.65 n=2.936
  [sim] Eval 26: s0=121 h0=2.4e+03 ss=542 n=2.94
  [sim] New best saved: RMSE=4.9548 MPa
  [sim] RMSE=4.955 MPa  MAPE=2.85%
  [sim] RMSE=4.955 < 5.0 MPa. Stopping.

[Tool 1] Stopped: RMSE=4.955 MPa
@R@[ReAct] Observation: COMPLETE: run_bayesian_optimization finished. Best parameters -- s0=120.6103 MPa, h0=2398.6739 MPa, ss=541.6495 MPa, n=2.9357. RMSE=4.9548 MPa, MAPE=2.846%. Algorithm: Bayesian Optimization (LHS + GP+EI). Please provide your physical interpretation of these results.@r@

@R@[ReAct] --- Iteration 3 ---@r@

@R@[ReAct] Thought: The calibrated slip parameters for SS316L austenitic stainless steel are as follows:

- **Initial Slip Resistance (s0 = 120.61 MPa):** This value indicates the yield stress of the material, suggesting that SS316L begins to plastically deform at a moderate stress level.
- **Initial Hardening Modulus (h0 = 2398.67 MPa):** A relatively high hardening modulus implies that the material exhibits significant strain hardening, which enhances its strength as it deforms.
- **Saturation Stress (ss = 541.65 MPa):** This parameter reflects the flow stress plateau, indicating the stress level at which the material's hardening rate decreases and approaches a steady state.
- **Power Law Exponent (n = 2.94):** A higher exponent value suggests a sharp transition from elastic to plastic behavior, indicating that SS316L has a pronounced yield point.

Overall, these parameters describe a material that has a moderate yield stress, strong strain hardening capabilities, and a distinct transition from elastic to plastic deformation, which are characteristic of SS316L's mechanical behavior.@r@

=================================================================
@R@[ReAct] CALIBRATION COMPLETE@r@

@R@[ReAct] Final Interpretation:
The calibrated slip parameters for SS316L austenitic stainless steel are as follows:

- **Initial Slip Resistance (s0 = 120.61 MPa):** This value indicates the yield stress of the material, suggesting that SS316L begins to plastically deform at a moderate stress level.
- **Initial Hardening Modulus (h0 = 2398.67 MPa):** A relatively high hardening modulus implies that the material exhibits significant strain hardening, which enhances its strength as it deforms.
- **Saturation Stress (ss = 541.65 MPa):** This parameter reflects the flow stress plateau, indicating the stress level at which the material's hardening rate decreases and approaches a steady state.
- **Power Law Exponent (n = 2.94):** A higher exponent value suggests a sharp transition from elastic to plastic behavior, indicating that SS316L has a pronounced yield point.

Overall, these parameters describe a material that has a moderate yield stress, strong strain hardening capabilities, and a distinct transition from elastic to plastic deformation, which are characteristic of SS316L's mechanical behavior.@r@
=================================================================

@R@[ReAct] Generating plots...@r@
Loaded best params from best_params.json: RMSE=4.9548 MPa
Total iterations loaded: 26
Running validation simulation with best parameters...
Validation RMSE: 4.9548 MPa
Curves loaded: 26
Saved: S3_best_fit.png
Saved: S3_convergence.png
/home/samoalfred/candi/plasticity/applications/crystalPlasticity/fcc/
SS316L_Semi_LLM_React_MicroGen3/plot_results.py:215: UserWarning: This figure includes Axes
that are not compatible with tight_layout, so results might be incorrect.
  fig3.tight_layout()
Saved: S3_sensitivity.png
Saved: S3_all_curves.png

All plots saved to: /home/samoalfred/candi/plasticity/applications/crystalPlasticity/fcc/SS316L_Semi_LLM_React_MicroGen3/workdir

@R@Final Summary:
  Best RMSE : 4.9548 MPa
  Best MAPE : 2.8464 %
  s0        : 120.6103 MPa
  h0        : 2398.6739 MPa
  ss        : 541.6495 MPa
  n         : 2.9357@r@
samoalfred@SAMOALFRED:~/candi/plasticity/applications/crystalPlasticity/fcc/SS316L_Semi_LLM_React_MicroGen3$
\end{lstlisting}

\section*{Acknowledgments}
The authors would like to acknowledge the Air Force Office of Scientific Research, National Science Portal Pilot program titled `Center for Scientific Machine Learning for Material Science' (Grant No. FA9550-23-1-0725) for supporting this work.

\section*{Code and Data Availability}
The agent harness (system prompt, tool schemas, dispatcher, and iteration loop), the per-case user queries, the microstructure-generation and post-processing scripts, and the calibration/target data supporting the four case studies are available at \url{https://github.com/samoalfred/harness-cp-agents}. The underlying LLM (GPT-4o) is accessed through the OpenAI API \cite{ref47}. Data from \cite{yaghoobi2025effects} and \cite{yaghoobi2022prisms} were accessed from Materials Commons at \url{https://materialscommons.org/public/datasets/193/overview}.

\section*{CRediT authorship contribution statement}
\textbf{Samuel Alfred:} Writing--original draft, Writing--review \& editing, Software, Methodology, Investigation, Data curation. \textbf{Abhishek Kumar:} Writing--review \& editing, Software. \textbf{Veera Sundararaghavan:} Writing--review \& editing, Writing--original draft, Supervision, Project administration, Funding acquisition, Conceptualization.

\section*{Declaration of generative AI and AI-assisted technologies in the manuscript preparation process}
During the preparation of this work, the authors used ChatGPT and Claude, large language models developed by OpenAI and Anthropic, respectively, in order to assist with language and grammar refinement. After using these tools, the authors reviewed and edited the content as needed and take full responsibility for the content of the published article. The scientific content, technical interpretations, and conclusions are solely the responsibility of the authors.

\section*{Declaration of competing interest}
The authors declare that they have no known competing financial interests or personal relationships that could have appeared to influence the work reported in this study.

\bibliographystyle{elsarticle-num-names}
\bibliography{cas-refs}

@article{ref1,
  author  = {Roters, F. and Eisenlohr, P. and Hantcherli, L. and others},
  title   = {Overview of constitutive laws, kinematics, homogenization and multiscale methods in crystal plasticity finite-element modeling: Theory, experiments, applications},
  journal = {Acta Materialia},
  volume  = {58},
  pages   = {1152--1211},
  year    = {2010},
  doi     = {10.1016/j.actamat.2009.10.058}
}

@article{ref2,
  author  = {Bhattacharya, K.},
  title   = {Multiscale modeling of materials and neural operators},
  journal = {MRS Bulletin},
  volume  = {51},
  pages   = {1--11},
  year    = {2026},
  doi     = {10.1557/s43577-026-01117-8}
}

@article{ref3,
  author  = {Asaro, R. J.},
  title   = {Micromechanics of Crystals and Polycrystals},
  journal = {Advances in Applied Mechanics},
  volume  = {23},
  pages   = {1--115},
  year    = {1983}
}

@article{ref4,
  title={An analysis of nonuniform and localized deformation in ductile single crystals},
  author={Peirce, D and Asaro, RJ and Needleman, A},
  journal={Acta metallurgica},
  volume={30},
  number={6},
  pages={1087--1119},
  year={1982},
  publisher={Elsevier}
}

@article{ref5,
  author  = {Eisenlohr, P. and Roters, F.},
  title   = {Selecting a set of discrete orientations for accurate texture reconstruction},
  journal = {Computational Materials Science},
  volume  = {42},
  pages   = {670--678},
  year    = {2008},
  doi     = {10.1016/j.commatsci.2007.09.015}
}

@article{ref6,
  author  = {Raabe, D. and Roters, F.},
  title   = {Using texture components in crystal plasticity finite element simulations},
  journal = {International Journal of Plasticity},
  volume  = {20},
  pages   = {339--361},
  year    = {2004},
  doi     = {10.1016/S0749-6419(03)00092-5}
}

@book{ref7,
  title={Texture and anisotropy: preferred orientations in polycrystals and their effect on materials properties},
  author={Kocks, U Fred and Tom{\'e}, Carlos Norberto and Wenk, H-R},
  year={2000},
  publisher={Cambridge university press}
}

@article{ref8,
  title={Physics and phenomenology of strain hardening: the FCC case},
  author={Kocks, UF and Mecking, Heinrich},
  journal={Progress in materials science},
  volume={48},
  number={3},
  pages={171--273},
  year={2003},
  publisher={Elsevier}
}

@article{ref9,
  author  = {Knezevic, M. and Savage, D. J.},
  title   = {A high-performance computational framework for fast crystal plasticity simulations},
  journal = {Computational Materials Science},
  volume  = {83},
  pages   = {101--106},
  year    = {2014},
  doi     = {10.1016/j.commatsci.2013.11.012}
}

@article{ref10,
  author  = {Tom\'e, C. and Canova, G. R. and Kocks, U. F. and others},
  title   = {The relation between macroscopic and microscopic strain hardening in {F.C.C.} polycrystals},
  journal = {Acta Metallurgica},
  volume  = {32},
  pages   = {1637--1653},
  year    = {1984},
  doi     = {10.1016/0001-6160(84)90222-0}
}

@article{ref11,
  author  = {Gupta, T. and Zaki, M. and Krishnan, N. M. A. and Mausam},
  title   = {{MatSciBERT}: A materials domain language model for text mining and information extraction},
  journal = {npj Computational Materials},
  volume  = {8},
  pages   = {1--11},
  year    = {2022},
  doi     = {10.1038/s41524-022-00784-w}
}

@article{ref12,
  author  = {Kononova, O. and He, T. and Huo, H. and others},
  title   = {Opportunities and challenges of text mining in materials research},
  journal = {iScience},
  volume  = {24},
  pages   = {102155},
  year    = {2021},
  doi     = {10.1016/j.isci.2021.102155}
}

@article{ref13,
  author  = {Ansari, M. and Moosavi, S. M.},
  title   = {Agent-based learning of materials datasets from the scientific literature},
  journal = {Digital Discovery},
  volume  = {3},
  pages   = {2607--2617},
  year    = {2024},
  doi     = {10.1039/d4dd00252k}
}

@article{ref14,
  author  = {Dagdelen, J. and Dunn, A. and Lee, S. and others},
  title   = {Structured information extraction from scientific text with large language models},
  journal = {Nature Communications},
  volume  = {15},
  year    = {2024},
  doi     = {10.1038/s41467-024-45563-x}
}

@article{ref15,
  author  = {Polak, M. P. and Morgan, D.},
  title   = {Extracting accurate materials data from research papers with conversational language models and prompt engineering},
  journal = {Nature Communications},
  volume  = {15},
  pages   = {1--11},
  year    = {2024},
  doi     = {10.1038/s41467-024-45914-8}
}

@article{ref16,
  author  = {Gupta, S. and Mahmood, A. and Shetty, P. and others},
  title   = {Data extraction from polymer literature using large language models},
  journal = {Communications Materials},
  volume  = {5},
  pages   = {1--11},
  year    = {2024},
  doi     = {10.1038/s43246-024-00708-9}
}

@article{ref17,
  author  = {Ghosh, S. and Tewari, A.},
  title   = {{LLM}-based {AI} agents for automated extraction of material properties and structural features},
  journal = {Computational Materials Science},
  volume  = {265},
  year    = {2026},
  doi     = {10.1016/j.commatsci.2026.114521}
}

@misc{ref18,
  author = {Shi, Y. and Gong, Y. and Su, Y. and others},
  title  = {Aethorix v1.0: An Integrated Scientific {AI} Agent for Scalable Inorganic Materials Innovation and Industrial Implementation},
  note   = {Preprint, pp. 1--15},
  year   = {2025}
}

@article{ref19,
  author  = {Zeng, T. and Badrinarayanan, S. and Ock, J. and others},
  title   = {{LLM}-guided chemical process optimization with a multi-agent approach},
  journal = {Machine Learning: Science and Technology},
  volume  = {6},
  year    = {2025},
  doi     = {10.1088/2632-2153/ae2382}
}

@article{ref20,
  author  = {Xiaorui, L. and Yuhao, Z. and Leiqi, W. and others},
  title   = {{LLMs} driven fusion {AI-AD} system for mechanical design: From understanding to generation},
  journal = {Advanced Engineering Informatics},
  volume  = {68},
  pages   = {103745},
  year    = {2025},
  doi     = {10.1016/j.aei.2025.103745}
}

@misc{ref21,
  author = {Guo, B. and Li, W. and Liu, X. and others},
  title  = {A Multidisciplinary Design and Optimization ({MDO}) Agent Driven by Large Language Models},
  note   = {Preprint},
  year   = {2025}
}

@article{ref22,
  author  = {Guo, J. and Park, C. and Kam, W. and Hughes, T. J. R.},
  title   = {Large language model-empowered next-generation computer-aided engineering},
  journal = {Computer Methods in Applied Mechanics and Engineering},
  volume  = {450},
  year    = {2026}
}

@article{ref23,
  author  = {Deotale, R. and Srinivasan, A. and Golestanian, M. and others},
  title   = {{ALL-FEM}: Agentic Large Language Models fine-tuned for finite element methods},
  journal = {Computer Methods in Applied Mechanics and Engineering},
  volume  = {457},
  pages   = {118985},
  year    = {2026},
  doi     = {10.1016/j.cma.2026.118985}
}

@misc{ref24,
  author = {Lin, H. and Zhang, W. and Xu, W. and others},
  title  = {{TO-Master}: an {LLM}-agent framework for automated topology optimization},
  note   = {Preprint, pp. 1--29},
  year   = {2026}
}

@misc{ref25,
  author = {Lin, C. and Hai, Y. and He, Y. and others},
  title  = {{CAX-Agent}: A Lightweight Agent Harness for Reliable {APDL} Automation},
  note   = {Preprint},
  year   = {2026}
}

@misc{ref26,
  author = {Wilke, D. N.},
  title  = {From Perception to Autonomous Computational Modeling: A Multi-Agent Approach},
  note   = {Preprint, pp. 1--34},
  year   = {2026}
}

@article{ref27,
  author  = {Alfred, S. O. and Sundararaghavan, V.},
  title   = {From data to theory: Autonomous large language model agents for materials science},
  journal = {Computational Materials Science},
  volume  = {272},
  pages   = {114862},
  year    = {2026},
  doi     = {10.1016/j.commatsci.2026.114862}
}

@misc{ref28,
  author = {Zhang, T. and Liu, Z. and Xin, Y. and Jiao, Y.},
  title  = {{MooseAgent}: A {LLM} Based Multi-agent Framework for Automating {MOOSE} Simulation},
  note   = {Preprint, pp. 1--11},
  year   = {2025}
}

@misc{ref29,
  author = {Wang, Z. and Huang, H. and Zhao, H. and others},
  title  = {{DREAMS}: Density Functional Theory Based Research Engine for Agentic Materials Simulation},
  note   = {Preprint},
  year   = {2025}
}

@article{ref30,
  author  = {Shi, Z. and Xin, C. and Huo, T. and others},
  title   = {A fine-tuned large language model based molecular dynamics agent for code generation to obtain material thermodynamic parameters},
  journal = {Scientific Reports},
  volume  = {15},
  pages   = {1--11},
  year    = {2025},
  doi     = {10.1038/s41598-025-92337-6}
}

@article{ref31,
  author  = {Wang, X. and Zeng, Q. and Xu, D. and others},
  title   = {Accelerating materials discovery via {AI-Agent} integration of large language models and simulation tools},
  journal = {Journal of Materials Informatics},
  volume  = {6},
  pages   = {1--15},
  year    = {2026},
  doi     = {10.20517/jmi.2025.69}
}

@article{ref32,
  author  = {Chaudhari, A. and Ock, J. and Barati Farimani, A.},
  title   = {Modular large language model agents for multi-task computational materials science},
  journal = {Communications Materials},
  volume  = {7},
  year    = {2026},
  doi     = {10.1038/s43246-025-00994-x}
}

@misc{ref33,
  author = {Yue, L. and Somasekharan, N. and Zhang, T. and others},
  title  = {{Foam-Agent} 2.0: An End-to-End Composable Multi-Agent Framework for Automating {CFD} Simulation in {OpenFOAM}},
  note   = {Preprint, pp. 1--28},
  year   = {2025}
}

@misc{ref34,
  author = {Pham, T. D. and Tummalapalli, H. and Bhuiyan, F. H. and others},
  title  = {Multi-Agent Orchestration for High-Throughput Materials Screening on a Leadership-Class System},
  note   = {arXiv:2604.07681, pp. 1--13},
  year   = {2026}
}

@article{ref35,
  author  = {Yang, J. and Kobayashi, Y. and Demura, M.},
  title   = {{AI} agents for automating materials research: a case study of crystal plasticity simulations},
  journal = {Science and Technology of Advanced Materials: Methods},
  volume  = {6},
  year    = {2026},
  doi     = {10.1080/27660400.2026.2630445}
}

@inproceedings{ref36,
  author    = {Yao, S. and Zhao, J. and Yu, D. and others},
  title     = {{ReAct}: Synergizing Reasoning and Acting in Language Models},
  booktitle = {11th International Conference on Learning Representations (ICLR 2023)},
  pages     = {1--33},
  year      = {2023}
}

@misc{ref37,
  author = {Lin, J. and Liu, S. and Pan, C. and others},
  title  = {Agentic Harness Engineering: Observability-Driven Automatic Evolution of Coding-Agent Harnesses},
  note   = {Preprint, pp. 1--35},
  year   = {2026}
}

@misc{ref38,
  author = {Zhou, C. and Chai, H. and Chen, W. and others},
  title  = {Externalization in {LLM} Agents: A Unified Review of Memory, Skills, Protocols and Harness Engineering},
  note   = {Preprint, pp. 1--54},
  year   = {2026}
}

@misc{ref39,
  author = {Zhong, H. and Zhu, S.},
  title  = {{AI} Harness Engineering: A Runtime Substrate for Foundation-Model Software Agents},
  note   = {Preprint, pp. 1--16},
  year   = {2026}
}

@article{ref40,
  author  = {He, C. and Zhou, X. and Wang, D. and others},
  title   = {Harness Engineering for Language Agents: The Harness Layer as Control, Agency, and Runtime},
  journal = {Preprints},
  pages   = {0--20},
  year    = {2026},
  doi     = {10.20944/preprints202603.1756.v1}
}

@misc{ref41,
  author = {Hurst, A. and Lerer, A. and Goucher, A. P. and Perelman, A. and Ramesh, A. and Clark, A. and others},
  title  = {{GPT-4o} system card},
  note   = {OpenAI},
  year   = {2024}
}

@misc{ref42,
  author = {Li, H. and Leung, J. and Shen, Z.},
  title  = {Towards Goal-oriented Prompt Engineering for Large Language Models: A Survey},
  note   = {Preprint},
  year   = {2024}
}

@article{ref43,
  author  = {Jeong, J. and Sundararaghavan, V.},
  title   = {{ELAS3D-Xtal}: An {OpenMP}-accelerated crystal elasticity solver with automated experiment-driven microstructure generation},
  journal = {Computational Materials Science},
  volume  = {269},
  pages   = {114734},
  year    = {2026},
  doi     = {10.1016/j.commatsci.2026.114734}
}

@article{ref44,
  author  = {Yaghoobi, M. and Ganesan, S. and Sundar, S. and others},
  title   = {{PRISMS-Plasticity}: An open-source crystal plasticity finite element software},
  journal = {Computational Materials Science},
  volume  = {169},
  year    = {2019},
  doi     = {10.1016/j.commatsci.2019.109078}
}

@article{ref45,
  title={A computational procedure for rate-independent crystal plasticity},
  author={Anand, Lallit and Kothari, Mrityunjay},
  journal={Journal of the Mechanics and Physics of Solids},
  volume={44},
  number={4},
  pages={525--558},
  year={1996},
  publisher={Elsevier}
}

@article{ref46,
  author  = {Bachmann, F. and Hielscher, R. and Schaeben, H.},
  title   = {Texture analysis with {MTEX} -- Free and open source software toolbox},
  journal = {Solid State Phenomena},
  volume  = {160},
  pages   = {63--68},
  year    = {2010},
  doi     = {10.4028/www.scientific.net/SSP.160.63}
}

@misc{ref47,
  author       = {{OpenAI}},
  title        = {Function calling},
  howpublished = {\url{https://developers.openai.com/api/docs/guides/function-calling}},
  note         = {Accessed 23 Jul 2026},
  year         = {2026}
}

@article{ref48,
  author  = {Douglas, R. and Beard, W. and Barnard, N. and others},
  title   = {The influence of energy density on the low cycle fatigue behaviour of laser powder bed fused stainless steel 316L},
  journal = {International Journal of Fatigue},
  volume  = {181},
  pages   = {108123},
  year    = {2024},
  doi     = {10.1016/j.ijfatigue.2023.108123}
}

@article{tran2026single,
  title={Single-agent LLMs outperform multi-agent systems on multi-hop reasoning under equal thinking token budgets},
  author={Tran, Dat and Kiela, Douwe},
  journal={arXiv preprint arXiv:2604.02460},
  year={2026}
}

@article{bronkhorst1992polycrystalline,
  title={Polycrystalline plasticity and the evolution of crystallographic texture in FCC metals},
  author={Bronkhorst, Curt A and Kalidindi, Surya R and Anand, Lallit},
  journal={Philosophical Transactions: Physical Sciences and Engineering},
  pages={443--477},
  year={1992},
  publisher={JSTOR}
}

@article{yaghoobi2022prisms,
  title={Prisms-plasticity TM: an open-source rapid texture evolution analysis pipeline},
  author={Yaghoobi, Mohammadreza and Allison, John E and Sundararaghavan, Veera},
  journal={Integrating Materials and Manufacturing Innovation},
  volume={11},
  number={4},
  pages={479--496},
  year={2022},
  publisher={Springer}
}

@article{yaghoobi2025effects,
  title={Effects of Zn and Ca on the deformation texture evolution of Mg--Zn--Ca alloys at elevated temperatures},
  author={Yaghoobi, Mohammadreza and Berman, Tracy and Allison, John E},
  journal={Journal of Materials Research and Technology},
  volume={35},
  pages={37--50},
  year={2025},
  publisher={Elsevier}
}

\end{document}